%% file: iclr2027_conference.tex
\documentclass{article}
\usepackage{iclr2027_conference,times}
\input{math_commands.tex}

\usepackage{hyperref}
\usepackage{url}
\usepackage{booktabs}
\usepackage{graphicx}
\usepackage{amsmath,amssymb,amsthm}
\usepackage{stmaryrd}
\usepackage{multirow}
\usepackage[table]{xcolor}
\usepackage{enumitem}
\usepackage{fvextra}
\fvset{breaklines=true,breakanywhere=false,breakindent=1.5em,breaksymbolleft={},breaksymbolright={},breaksymbolsep=0pt,fontsize=\footnotesize}

\usepackage{algorithm}
\usepackage{algpseudocode}
\usepackage{subcaption}
\usepackage{tcolorbox}
\tcbuselibrary{breakable,skins}
\usepackage{placeins}

\newcommand{\meas}[1]{\ensuremath{#1}}
\newcommand{\rise}{\textsc{RISE}}
\newcommand{\secbandit}{\textsc{SEC}}
\newcommand{\normset}[1]{[\![ #1 ]\!]}

\newsavebox{\tabboxsave}
\newcommand{\tabbox}[1]{%
  \sbox{\tabboxsave}{\small #1}%
  \ifdim\wd\tabboxsave>\linewidth
    \resizebox{\linewidth}{!}{\usebox{\tabboxsave}}%
  \else
    \usebox{\tabboxsave}%
  \fi}
\newcommand{\best}[1]{\textbf{#1}}
\newcommand{\second}[1]{\underline{#1}}
\newcommand{\riserow}{\rowcolor{blue!8}}

\newtcolorbox{promptbox}[1]{
  colback=gray!3,colframe=gray!45,
  fonttitle=\bfseries\small,coltitle=white,colbacktitle=gray!60,
  boxrule=0.4pt,arc=2pt,left=4pt,right=4pt,top=3pt,bottom=3pt,
  breakable,title=#1}

\newtheorem{proposition}{Proposition}

\title{Interactive Structured Environment Scaling\\
\scalebox{0.95}[1.0]{for Large Language Model Agentic Reasoning}}

\author{Peng Yu$^{1}$ \quad Luoyi Fu$^{1}$ \quad Xinbing Wang$^{1}$ \quad Ying Wen$^{2}$\\
\small $^{1}$Department of Computer Science and Engineering, Shanghai Jiao Tong University\\
\small $^{2}$School of Artificial Intelligence and John Hopcroft Center, Shanghai Jiao Tong University}

\begin{document}
\maketitle
\lhead{Preprint.}
\addtocontents{toc}{\protect\setcounter{tocdepth}{-1}}

\begin{abstract}
Reinforcement learning via environmental interaction has become a cornerstone post-training paradigm for evolving large language models into autonomous agents capable of multi-turn tool use and knowledge exploration. Because the training environment defines what the policy can observe and how its actions are verified, the environment's structure directly bounds the capabilities an agent can acquire. Yet existing structured and retrieval environments predominantly rely on upstream pre-filtering---extracting a compact per-question subgraph around the gold answer or retrieving a fixed passage list before the policy acts. We show that such pre-filtering introduces a fundamental measurement confound: because per-question subgraphs are constructed to enclose ground-truth paths, they compress the candidate space by orders of magnitude and leak answer locations through local topology, allowing a question-blind degree heuristic to surface gold answers far above chance. To restore path selection, candidate pruning, and optimal stopping as genuine policy decisions, we propose \rise{} (Inte\textbf{R}act\textbf{I}ve \textbf{S}tructured \textbf{E}nvironment), transitioning structured reinforcement learning from static in-context subgraphs to full-scale interactive databases along two coupled axes. First, \emph{Decision-Space Scaling} exposes the unpruned knowledge graph through a five-stage graph-agnostic typed intent adapter under a bounded call budget and a multiplicative grounding reward gate. Second, \emph{Learning-Frontier Scaling} overcomes the severe within-group variance collapse of full-graph exploration under group-normalized objectives by uniting model-independent structural tier progression with a self-evolving competence-frontier bandit. Across nineteen benchmarks and three model backbones, \rise{} substantially advances in-domain and unseen-schema knowledge-graph reasoning, transfers zero-shot to open-domain passage retrieval, and significantly enhances out-of-domain mathematical and logical deduction.
\end{abstract}

\section{Introduction}
\label{sec:intro}

Fine-tuning large language models (LLMs) with reinforcement learning (RL) and verifiable feedback has driven a paradigm shift in post-training, eliciting deliberative capabilities such as hypothesis testing, self-reflection, backtracking, and multi-step planning \citep{jaech2024openai,guo2025deepseekr1,team2025kimi,comanici2025gemini}. By optimizing policies directly against outcome verifiers, RL enables models to internalize reasoning strategies that are difficult to instil through passive imitation alone \citep{shao2024deepseekmath,zeng2025simplerl,hu2025openreasoner}. While early breakthroughs centered on symbolic domains such as mathematics, logic, and code generation \citep{cobbe2021gsm8k,lightman2023lets,xie2025logicrl,chen2025r1code}, this paradigm is rapidly expanding to interactive agentic settings, empowering LLMs to query search engines, navigate databases, and orchestrate external tools over multi-turn horizons \citep{yao2023react,schick2023toolformer,jin2025searchr1,zheng2025deepresearcher,li2025webthinker,qian2025toolrl,xue2026simpletir}.

Despite rapid progress in RL optimization algorithms \citep{schulman2017ppo,shao2024deepseekmath,hu2025reinforce,yu2025dapo,zheng2025gspo}, the foundational role of the \emph{training environment} itself has been comparatively overlooked \citep{fang2025general,yu2026sie}. In interactive RL, the environment governs what the policy observes, which actions are admissible, and when an episode terminates, directly bounding what decision-making capabilities a policy can acquire (see Appendix~\ref{app:related} for a detailed review of related work). Existing environments face a persistent trade-off across these dimensions. Internalized-rule environments (e.g., mathematics and code) provide clean verifiability, but rely on costly expert annotations and do not train agents to gather missing evidence under partial observability \citep{hendrycks2021math,lightman2023lets}. Conversely, externalized-rule game and embodied environments support multi-turn exploration \citep{carta2023grounding,tan2024true,wen2024reinforcing,zhang2025leveraging}, yet the learned skills often overfit to specialized mechanics and fail to transfer to general reasoning. Large-scale structured knowledge bases offer an ideal foundation to bridge this gap \citep{codd1970relational,bollacker2008freebase,lacombe2025reasoning,wang2025graphr1,yu2026kalm,yu2026sie}: their expansive schemas support automated scaling, their multi-hop paths mirror compositional deduction, and their explicit facts enable exact rule-based verification.

However, for a structured environment to genuinely incentivize transferable \emph{agentic search and reasoning} rather than localized reading comprehension, it must simultaneously satisfy three core criteria (formalized in Appendix~\ref{app:prelim}): \textbf{(1) Open Decision Space}, where sequential path selection, candidate pruning, and optimal stopping governance remain open decisions for the policy; \textbf{(2) Tool-Agnostic Action Interface}, where query formulation is decoupled from backend-specific query syntax so that learned search habits transfer across heterogeneous tools; and \textbf{(3) Active Learning Frontier}, where the training schedule sustains non-zero policy-gradient variance as the decision space scales. Yet existing structured and retrieval environments---whether injecting serialized subgraphs into the prompt \citep{luo2024rog,he2024gretriever,yu2026kalm,yu2026sie}, walking pre-extracted question-specific subgraphs \citep{sun2024tog,chen2024pog,lin2025kgr1,xu2026graphwalker}, or retrieving fixed top-$k$ passages \citep{jin2025searchr1,chen2025research,song2025r1searcher,sun2025zerosearch}---violate the first criterion by design: \emph{they pre-filter the candidate space before the policy takes a single step}.

\begin{figure}[t]
\centering
\includegraphics[width=0.95\linewidth]{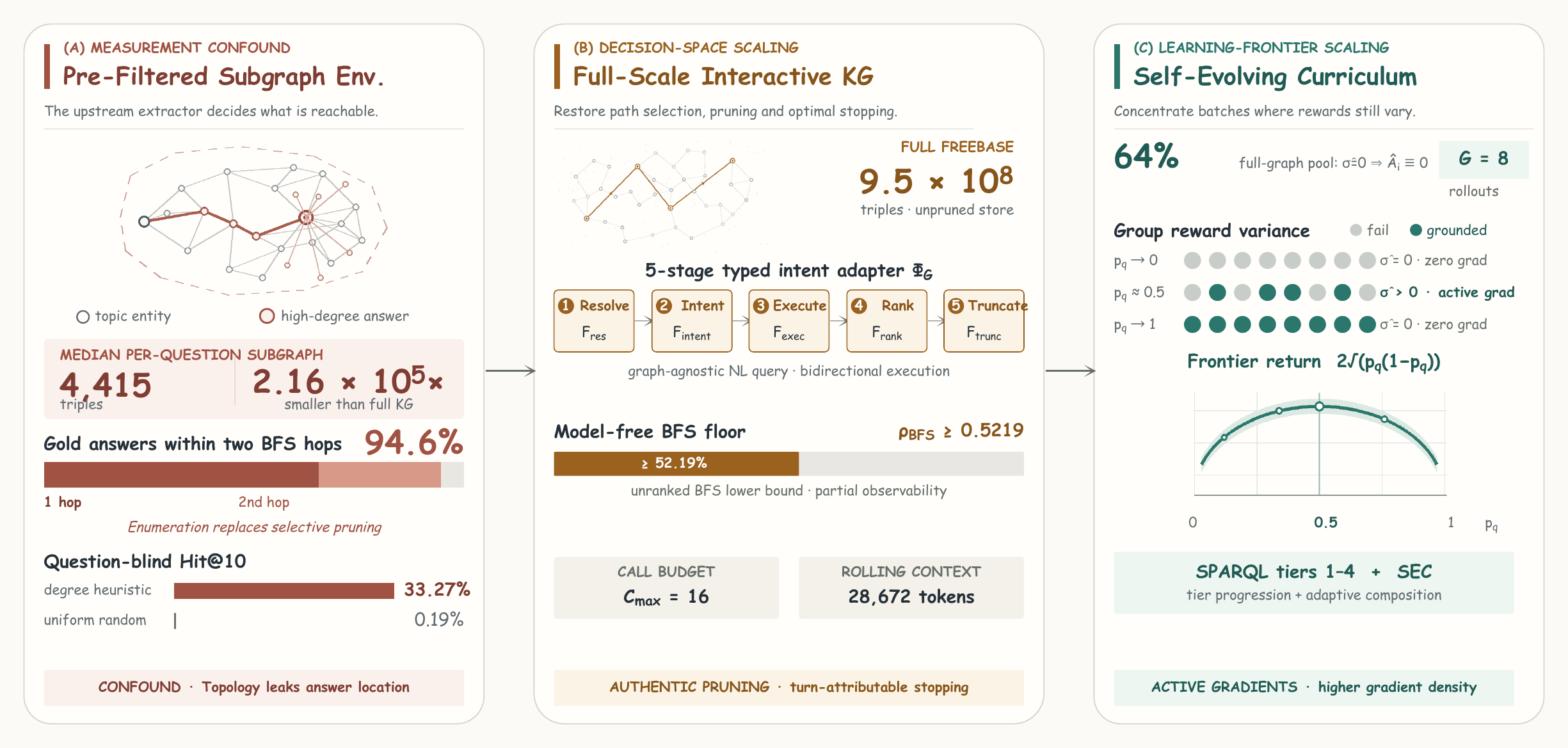}
\vspace{-4pt}
\caption{Motivation and the two scaling axes of \rise{}. \textbf{(A)} Pre-filtered subgraph environments compress the candidate space by $2.16\times 10^5\times$ and leak answer locations through node degree ($33.27\%$ question-blind Hit@10 vs.\ $0.19\%$ uniform random). \textbf{(B)} \emph{Decision-Space Scaling} exposes the full $9.5\times 10^8$-triple store through a five-stage graph-agnostic typed intent adapter $\Phi_{\mathcal{G}}$ under a bounded call budget ($C_{\max}=16$) and rolling context ($28{,}672$ tokens). \textbf{(C)} \emph{Learning-Frontier Scaling} overcomes the $64\%$ zero-variance bottleneck of full-graph group-normalized RL ($G=8$) by uniting SPARQL structural tiers 1--4 with the Self-Evolving Curriculum (\secbandit{}) bandit ($p_q \approx 0.5$; see Appendix~\ref{app:prelim}).}
\label{fig:paradigm-scales}
\vspace{-6pt}
\end{figure}

Treating such pre-filtered contexts as search environments introduces a severe \emph{measurement confound} before it imposes any capability ceiling. We quantify this confound on the per-question WebQSP subgraphs released by a widely used RL knowledge-graph agent \citep{lin2025kgr1}, auditing them exactly as published (Appendix~\ref{app:prelim}). Three findings stand out. First, the candidate space is compressed by five orders of magnitude, with the median subgraph holding only $\meas{4{,}415}$ triples over $\meas{1{,}638}$ entities ($\meas{2.16\times 10^5}$ times smaller than full Freebase). Second, $\meas{94.6\%}$ of gold answers reside within two hops of a topic entity ($\meas{65.5\%}$ within one hop), allowing naive breadth-first enumeration to substitute for selective planning. Third, and most critically, because each subgraph is extracted to enclose the ground-truth path, its local connectivity directly leaks the answer's location: a question-blind heuristic that never reads the question and simply ranks nodes by in-subgraph degree surfaces a gold answer in the top-10 for $\meas{33.27\%}$ of questions, against $\meas{0.19\%}$ for a uniform random baseline (Figure~\ref{fig:paradigm-scales}A). When pre-extracted topology already points at the answer and guarantees reachability by construction ($\rho \equiv 1$), accuracy differences cannot be reliably attributed to a learned search policy.

To transform reachability from a construction artifact into a learnable policy outcome, we transition the training environment from static pre-extracted subgraphs to full-scale interactive databases and propose \rise{} (Inte\textbf{R}act\textbf{I}ve \textbf{S}tructured \textbf{E}nvironment). As illustrated in Figure~\ref{fig:paradigm-scales}, \rise{} scales agentic RL along two coupled dimensions. First, \emph{Decision-Space Scaling} (Figure~\ref{fig:paradigm-scales}B) exposes the unpruned Freebase store ($\meas{9.5\times 10^8}$ triples, where a model-free breadth-first probe establishes a reachability floor of $\rho_{\text{BFS}} \ge \meas{0.5219}$) under a finite call budget ($C_{\max}=16$) and a rolling context window ($28{,}672$ tokens), restoring candidate pruning and stopping control as authentic, turn-attributable policy decisions. Rather than binding actions to raw SPARQL syntax \citep{chen2025knowcoder,song2026ispkgr} or schema-specific primitives \citep{lin2025kgr1,gong2026tempr1}, \rise{} interposes a deterministic five-stage \emph{graph-agnostic typed intent adapter} $\Phi_{\mathcal{G}}$ that executes natural-language queries symmetrically across outgoing and incoming edges ($\meas{48.6\%}$ of candidate relations exist solely as incoming edges), paired with a multiplicative grounding reward gate and an isomorphic protocol cold start (Figure~\ref{fig:framework}). Because the policy interacts strictly through natural-language intents and structured receipts, it transfers zero-shot to unseen schemas, temporal knowledge graphs \citep{chen2024multitq,gong2026tempr1}, and unstructured passage retrievers \citep{yan2026eog}, evaluated unconditionally on full official splits.

Second, \emph{Learning-Frontier Scaling} (Figure~\ref{fig:paradigm-scales}C) addresses the optimization bottleneck that decision-space enlargement exposes: within-group variance collapse under group-normalized RL \citep{shao2024deepseekmath}. When advantages are standardized across $G=8$ rollouts sampled for the \emph{same} question, any question on which all rollouts earn identical rewards---occurring both when an easy question is already mastered ($p_q \to 1$) and when full-graph exploration fails completely on a hard or unreachable question ($p_q \to 0$)---yields zero within-group variance ($\hat{\sigma}=0$) and contributes zero gradient (Proposition~\ref{prop:variance}; affecting $\meas{64\%}$ of our training pool). Because zero variance is a per-question property, mixing mastered and unreachable questions in a batch still yields zero gradient even when the batch mean pass rate is $0.5$. We resolve this bottleneck across four deterministic structural tiers ($d(q) \in \{1,2,3,4\}$ parsed from gold SPARQL topology) at two levels: macroscopically, progressing from single-hop lookups to multi-constraint compositions aligns training direction with the outward expansion of the policy's reachable horizon \citep{bengio2009curriculum,kumar2010selfpaced}; microscopically, we adapt the self-evolving curriculum (\secbandit{}) scheduler of \citet{chen2025sec}---in the spirit of prioritized replay and bandit scheduling \citep{schaul2016prioritized,auer2002ucb}---by anchoring its return on the group-level Bernoulli standard deviation $2\sqrt{p_q(1-p_q)}$ to dynamically concentrate batches on the moving competence frontier ($p_q \approx 0.5$).

We instantiate \rise{} on Freebase \citep{bollacker2008freebase} and evaluate it across nineteen benchmarks spanning three capability families and three model backbones, making three main contributions:
\begin{itemize}[leftmargin=1.5em,topsep=2pt,itemsep=2pt]
\item We empirically diagnose the topological leakage confound of pre-filtered subgraph environments ($33.27\%$ question-blind Hit@10), and formalize \rise{} over an unpruned knowledge base so that path selection, candidate pruning, and optimal stopping become turn-attributable decisions.
\item We design a five-stage graph-agnostic typed intent adapter with a multiplicative grounding gate, prove within-group variance collapse under group normalization (Proposition~\ref{prop:variance}), and operationalize \emph{Learning-Frontier Scaling} by uniting SPARQL structural tiering with an online variance bandit (\secbandit{}).
\item Across nineteen benchmarks on Qwen2.5-7B, Qwen2.5-14B, and Llama-3.1-8B, we show that \rise{} outperforms specialized baselines on in-domain and unseen-schema KG reasoning, transfers zero-shot to passage retrieval above domain-trained models, and generalizes to tool-free mathematical and logical deduction, while isolating the effects of cold start, curriculum direction, and batch composition.
\end{itemize}

\section{The \texorpdfstring{\rise{}}{RISE} Framework}
\label{sec:method}

\begin{figure}[t]
\centering
\includegraphics[width=0.95\linewidth]{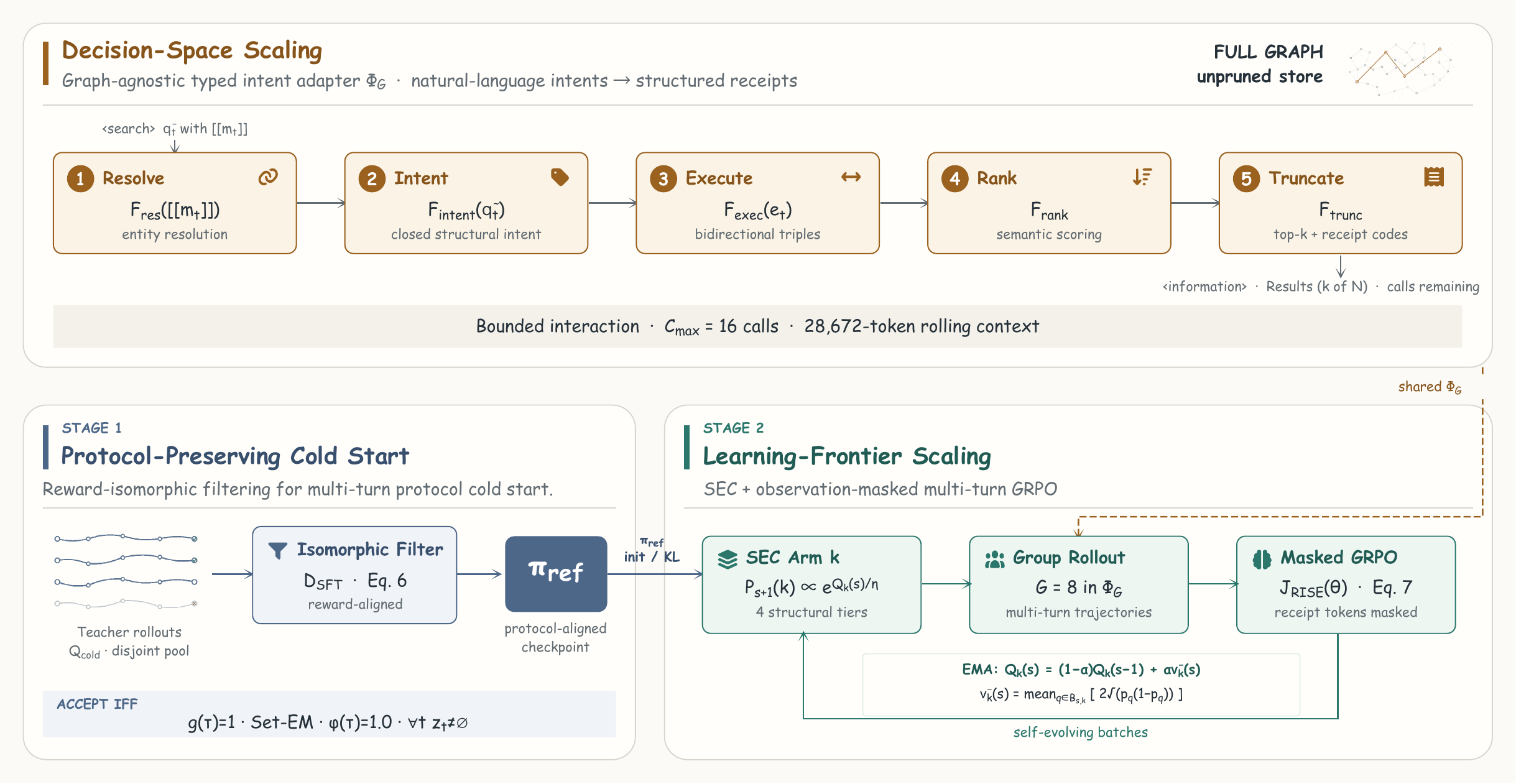}
\vspace{-4pt}
\caption{Overview of the \rise{} framework. \textbf{Top:} \emph{Decision-Space Scaling} interposes a five-stage graph-agnostic typed intent adapter $\Phi_{\mathcal{G}}$ over the full unpruned graph ($9.5\times 10^8$ triples), translating natural-language \texttt{<search>} intents into bidirectional traversals and structured \texttt{<information>} receipts under a bounded call budget ($C_{\max}=16$) and rolling context ($28{,}672$ tokens). \textbf{Bottom Left:} Stage 1 filters teacher rollouts on a strictly disjoint pool $\mathcal{Q}_{\text{cold}}$ using the exact grounded outcome criterion of the RL reward (Eq.~\ref{eq:sft-corpus}). \textbf{Bottom Right:} Stage 2 performs \emph{Learning-Frontier Scaling} by coupling observation-masked multi-turn GRPO (Eq.~\ref{eq:grpo-objective}) with the Self-Evolving Curriculum (\secbandit{}) bandit across four structural tiers (Eq.~\ref{eq:sec-bandit}).}
\label{fig:framework}
\vspace{-6pt}
\end{figure}

This section presents the \rise{} framework (Figure~\ref{fig:framework}), comprising three components: \emph{Decision-Space Scaling}, realized by a budgeted interactive POMDP formulation and a five-stage graph-agnostic typed intent adapter (Sections~\ref{sec:mdp}--\ref{sec:env}); a two-stage training pipeline combining a protocol-preserving cold start with observation-masked multi-turn GRPO under a multiplicative grounding gate (Section~\ref{sec:reward-choice}); and \emph{Learning-Frontier Scaling}, realized by structural tier progression and a self-evolving variance bandit derived from group-normalized policy gradients (Section~\ref{sec:curriculum}).

\subsection{Budgeted Interactive Search over a Structured Environment}
\label{sec:mdp}

Learning to search is learning a sequence of decisions, not learning a query format. An environment carries that signal only if sequential path selection, candidate pruning, and stopping governance are genuinely open: once candidates are pre-enumerated and the answer enclosed upstream, the only task left is local evidence stitching.

We formulate multi-turn question answering over a knowledge graph $\mathcal{G} = (\mathcal{E}, \mathcal{R}, \mathcal{T})$ as a budget-constrained Partially Observable Markov Decision Process (POMDP), defined by the tuple $\mathcal{M} = (\mathcal{Q}, \mathcal{H}, \mathcal{A}, \mathcal{O}, \Phi_{\mathcal{G}}, R, C_{\max})$. Given an input question $q \in \mathcal{Q}$ (accompanied by surface-form topic entities when provided by the benchmark), an episode unfolds over $T \le C_{\max} + 1$ discrete interaction turns. At turn $t$, conditioned on the interaction history $h_{t-1} = (z_1, a_1, o_1, \dots, z_{t-1}, a_{t-1}, o_{t-1}) \in \mathcal{H}$ (with $h_0 = \emptyset$), the policy $\pi_\theta$ first generates an internal reasoning trace $z_t$ enclosed in \texttt{<think>...</think>} tags, followed by a discrete action $a_t \in \mathcal{A}$. The action space is the disjoint union $\mathcal{A} = \mathcal{A}_{\text{search}} \cup \mathcal{A}_{\text{answer}}$, where $a_t \in \mathcal{A}_{\text{search}}$ emits a natural-language query inside \texttt{<search>...</search>} and $a_t \in \mathcal{A}_{\text{answer}}$ emits a final prediction $\hat{A}(\tau)$ inside \texttt{<answer>...</answer>} (with an empty tag denoting explicit abstention).

Whenever $a_t \in \mathcal{A}_{\text{search}}$, the environment executes the deterministic adapter $\Phi_{\mathcal{G}}$ and appends a structured receipt $o_t = \Phi_{\mathcal{G}}(a_t, h_{t-1}) \in \mathcal{O}$ enclosed in \texttt{<information>...</information>} as a separate user turn, so environmental text never enters the policy's own generation stream. The probability of a trajectory $\tau = (z_1, a_1, o_1, \dots, z_T, a_T)$ factorizes autoregressively over the policy's own tokens subject to a hard call budget $C_{\max} = 16$, while a rolling window of $28{,}672$ tokens is maintained by left-truncating the oldest turns:
\begin{equation}
\pi_\theta(\tau \mid q) = \prod_{t=1}^{T} \pi_\theta(z_t, a_t \mid q, h_{t-1}), \quad \text{subject to} \quad C(\tau) = \sum_{t=1}^{T} \mathbb{I}[a_t \in \mathcal{A}_{\text{search}}] \le C_{\max}.
\label{eq:pomdp-factorization}
\end{equation}
The two budgets are deliberately orthogonal: the call budget counts external searches and the per-turn limit ($2{,}048$ tokens) counts internal reasoning tokens, so a policy can neither buy extra hops by thinking less nor lose hops by thinking more.

When the call budget is exhausted at turn $t$, the environment appends a notice to $o_t$ and grants one final answer-only turn restricted to $\mathcal{A}_{\text{answer}}$, keeping ``ran out of search budget'' and ``failed to emit an answer'' strictly separable at evaluation time. The decision process follows the directed interaction flow
\begin{equation}
q \xrightarrow{\pi_\theta} (z_1, a_1^{\text{search}}) \xrightarrow{\Phi_{\mathcal{G}}} o_1 \xrightarrow{\pi_\theta} \cdots \xrightarrow{\Phi_{\mathcal{G}}} o_{T-1} \xrightarrow{\pi_\theta} (z_T, a_T^{\text{answer}}) \implies \hat{A}(\tau),
\label{eq:interaction-graph}
\end{equation}
in which every transition through $\Phi_{\mathcal{G}}$ is an opportunity for the policy to revise its hypothesis or prune unpromising branches rather than follow a pre-extracted path.

Three structural consequences separate this formulation from pre-filtered environments. First, budget-bounded reachability $\rho(q; C_{\max})$ (formalized in Appendix~\ref{app:prelim}) is identically one by construction inside an answer-enclosed subgraph, whereas over the full graph it becomes an empirical quantity governed by the policy's own path selections. Second, candidate pruning becomes mandatory, since the policy must select which relation to traverse among all edges incident to an entity. Third, evaluation is conducted unconditionally over full official splits rather than reachable subsets, preserving a fixed denominator across checkpoints.

\subsection{Graph-Agnostic Typed Intent Adapter and Environment Contract}
\label{sec:env}

Section~\ref{sec:mdp} defines the sequential decisions a search policy must make, but if the action space directly encodes a database's native query syntax (such as SPARQL), the decision space becomes bound to that schema and cannot transfer across tools. We therefore restrict the policy's action vocabulary to natural language plus bracketed entity mentions, placing all schema-specific translation behind a deterministic adapter $\Phi_{\mathcal{G}}$ so that the exact same policy weights can drive a static knowledge graph, a temporal graph, or a text passage retriever without retraining.

\rise{} implements this decoupling via a four-layer architecture: (1) a full-scale Virtuoso SPARQL backend hosting Freebase; (2) a deterministic five-stage \emph{typed intent adapter} $\Phi_{\mathcal{G}}$; (3) a strict environment contract governing turn grammar, receipt schemas, and error codes; and (4) a trainer-facing multi-turn conversation interface shared identically between structured graph search and unstructured passage retrieval. Because the top conversation layer is invariant across backends, swapping the adapter is the only change required to evaluate a graph-trained policy on a passage retriever.

Given a search action $a_t = \text{\texttt{<search>}}\,\tilde{q}_t\,\text{\texttt{</search>}}$ containing query $\tilde{q}_t$ and bracketed mentions $\text{\texttt{[[}}m_t\text{\texttt{]]}}$, the adapter composes five stages, $\Phi_{\mathcal{G}} = \mathcal{F}_{\text{trunc}} \circ \mathcal{F}_{\text{rank}} \circ \mathcal{F}_{\text{exec}} \circ \mathcal{F}_{\text{intent}} \circ \mathcal{F}_{\text{res}}$:
\begin{enumerate}[leftmargin=1.6em,topsep=2pt,itemsep=1pt]
\item \textbf{Entity Resolution ($\mathcal{F}_{\text{res}}$).} Maps each surface form $m_t$ to a node $e_t\in\mathcal{E}$ through the episode's entity-binding table, seeded with the benchmark's topic entities and extended by every previously returned receipt; a form unseen in the table falls back to exact label matching in $\mathcal{G}$. Because the table is maintained per episode, homonymous labels are disambiguated by entities the episode has already grounded rather than by a static global prior.
\item \textbf{Intent Classification ($\mathcal{F}_{\text{intent}}$).} Maps $\tilde{q}_t$ to one of three closed-vocabulary structural tags, $\iota_t\in\{\text{\texttt{relation\_hop}},\text{\texttt{relation\_list}},\text{\texttt{entity\_info}}\}$, via deterministic surface cues (returning relational triples, deduplicated incident relation names, or entity descriptions and types).
\item \textbf{Direction-Symmetric Execution ($\mathcal{F}_{\text{exec}}$).} Auditing Freebase reveals that $\meas{48.6\%}$ of candidate relations around topic entities exist \emph{only} as incoming edges (Appendix~\ref{app:prelim}). An adapter that traversed outgoing edges alone would list such a relation during inspection and return empty results when queried, penalizing a valid reasoning step. Execution therefore retrieves the bidirectional one-hop neighborhood of $e_t$, inverting incoming edges into canonical $e_t$-centric candidates:
\begin{equation}
\mathcal{F}_{\text{exec}}(e_t) = \{(e_t, r, e') \mid (e_t, r, e') \in \mathcal{T}\} \;\cup\; \{(e_t, r^{-1}, e') \mid (e', r, e_t) \in \mathcal{T}\}.
\label{eq:bidirectional-exec}
\end{equation}
\item \textbf{Semantic Ranking ($\mathcal{F}_{\text{rank}}$).} Scores candidate triples in $\mathcal{F}_{\text{exec}}(e_t)$ by dense cosine similarity (\texttt{E5-base-v2}) against $\tilde{q}_t$ and retains those above a fixed similarity threshold, falling back to the entity's relation list when none qualifies.
\item \textbf{Bounded Truncation and Receipt Formatting ($\mathcal{F}_{\text{trunc}}$).} Emits the top $k$ of $N$ matching rows behind an explicit \texttt{Results (k of N)} header---keeping ``fact does not exist'' ($N=0$) distinguishable from ``fact truncated by row cap'' ($k<N$)---and closes with \texttt{Calls remaining}, capping rows at $k=100$ and relations at $50$ (Appendix~\ref{app:config}).
\end{enumerate}

Three contract rules guarantee that training and evaluation share identical environment semantics. First, the call budget $C_{\max}$ is accounted independently of the per-turn generation limit ($2{,}048$ tokens). Second, all execution anomalies map to a closed seven-element error vocabulary---\texttt{E\_NO\_ENTITY}, \texttt{E\_ENTITY\_NOT\_FOUND}, \texttt{E\_INTENT\_UNSUPPORTED}, \texttt{E\_NO\_RESULTS}, \texttt{E\_QUERY\_FAILED}, \texttt{E\_BUDGET\_EXCEEDED}, and \texttt{E\_MALFORMED} (Appendix~\ref{app:contract})---where transient backend timeouts (\texttt{E\_QUERY\_FAILED}) are never penalized as policy errors. Third, observation independence holds end to end: cold-start collection, on-policy GRPO rollouts, and benchmark evaluation invoke the exact same adapter binary and receipt template.

\subsection{Grounded Outcome Verification and Two-Stage Training}
\label{sec:reward-choice}

Both training stages ask the same question of a trajectory---did it reach a correct answer via evidence the environment can vouch for---preventing the RL phase from having to unlearn shortcuts accepted during warm-up.

\textbf{Grounded Outcome Verification.} Rewarding a completed trajectory $\tau$ with predicted answer $\hat{A}(\tau)$ against ground truth $A^*(q)$ combines a gated outcome term with an additive format term:
\begin{equation}
r(\tau) = g(\tau) \cdot \mathbb{I}\!\left[ \normset{\hat{A}(\tau)} = \normset{A^*(q)} \right] + \varphi(\tau),
\label{eq:reward}
\end{equation}
where $\normset{\cdot}$ normalizes aliases and whitespace at set level, $\varphi(\tau) \in \{1.0, 0.8, 0.2, 0.1, 0.0\}$ is the five-tier structural format score of Search-R1 \citep{jin2025searchr1}, and $g(\tau)\in\{0,1\}$ is a \emph{multiplicative grounding gate} requiring every predicted entity to appear in a receipt retrieved during the episode:
\begin{equation}
g(\tau) = \mathbb{I}\!\left[ \forall \hat{a} \in \normset{\hat{A}(\tau)}, \ \hat{a} \in \bigcup_{t=1}^{T-1} \normset{\mathcal{E}(o_t)} \right].
\label{eq:grounding-gate}
\end{equation}

The structural asymmetry between a multiplicative gate $g(\tau)$ and an additive format term $\varphi(\tau)$ is deliberate. Grounding must be multiplicative because an ungrounded correct answer on an unreachable question is a parametric guess, and rewarding it incentivizes the policy to bypass search and enumerate plausible entities instead. Format compliance, by contrast, must be additive: a multiplicative format gate would zero out the gradient whenever the final answer is wrong, depriving the policy of syntactic feedback during early exploration when accuracy is low and degenerate repetition loops \citep{holtzman2020degeneration} must be suppressed. Intermediate process rewards are excluded because under within-question group normalization, a bonus for visiting an intermediate hop alters the advantage scale asymmetrically across question difficulties.

\textbf{Protocol-Preserving Cold Start.} All training questions originate from the combined official training splits of WebQSP and CWQ ($34{,}215$ questions in total), which we partition via stratified sampling by dataset and difficulty tier into two \emph{strictly disjoint} pools: a cold-start pool $\mathcal{Q}_{\text{cold}}$ of $6{,}000$ questions and an online RL pool $\mathcal{Q}_{\text{RL}}$ of $28{,}215$ questions (enforced by a run-level split flag). Stage 1 fine-tunes the base model exclusively on teacher trajectories collected inside \rise{} over $\mathcal{Q}_{\text{cold}}$. Using Gemini-2.5-pro at temperature $0.7$ with at most two attempts per question, $4{,}006$ questions ($66.8\%$) yielded trajectories passing our two filtering invariants, giving $\lvert\mathcal{D}_{\text{SFT}}\rvert=4{,}006$: (1) \emph{reward-criterion isomorphism}, requiring $g(\tau)\cdot \mathbb{I}[\normset{\hat{A}(\tau)} = \normset{A^*(q)}] = 1$ and $\varphi(\tau) = 1.0$; and (2) \emph{per-turn reasoning completeness}, requiring a non-empty \texttt{<think>} block ($z_t \neq \emptyset$) at every turn $t$:
\begin{equation}
\begin{split}
\mathcal{D}_{\text{SFT}} = \Big\{ (q, \tau) \ \Big|\ & q \in \mathcal{Q}_{\text{cold}}, \ \tau \sim \pi_{\text{teacher}}(\cdot \mid q), \\
& g(\tau)\cdot \mathbb{I}\!\left[\normset{\hat{A}(\tau)} = \normset{A^*(q)}\right] = 1, \ \varphi(\tau) = 1.0, \ \textstyle\prod_{t=1}^{T} \mathbb{I}[z_t \neq \emptyset] = 1 \Big\}.
\end{split}
\label{eq:sft-corpus}
\end{equation}

\textbf{Observation-Masked Multi-Turn GRPO.} Stage 2 operates over the disjoint RL pool $\mathcal{Q}_{\text{RL}}$ ($28{,}215$ questions). At each step, we sample $G=8$ trajectories per question from $\pi_{\theta_{\text{old}}}$ and compute the group-normalized advantage $\hat{A}_i = (r(\tau_i) - \bar{r}) / \hat{\sigma}$, setting $\hat{A}_i = 0$ whenever $\hat{\sigma}=0$. Let $\mathcal{I}_i$ index the tokens of $\tau_i$ generated by the policy---all of $\{(z_t,a_t)\}_{t=1}^{T_i}$ but none of the environment receipts $\{o_t\}$---so that observation tokens contribute zero gradient:
\begin{equation}
\begin{aligned}
\mathcal{J}_{\text{RISE}}(\theta) = \mathbb{E}_{q \sim \mathcal{P}_{\text{sched}},\, \{\tau_i\}_{i=1}^G \sim \pi_{\theta_{\text{old}}}(\cdot \mid q)} \Bigg[ \frac{1}{G} \sum_{i=1}^{G} \frac{1}{\lvert\mathcal{I}_i\rvert} \sum_{m \in \mathcal{I}_i} \Big( \min\!\Big( \rho_{i,m}(\theta)\hat{A}_i, \\
\text{clip}\big(\rho_{i,m}(\theta), 1-\epsilon, 1+\epsilon\big)\hat{A}_i \Big) - \beta \mathbb{D}_{\text{KL}}(\pi_\theta \| \pi_{\text{ref}}) \Big) \Bigg],
\end{aligned}
\label{eq:grpo-objective}
\end{equation}
where $\rho_{i,m}(\theta) = \frac{\pi_\theta(\tau_{i,m} \mid q, \tau_{i,<m})}{\pi_{\theta_{\text{old}}}(\tau_{i,m} \mid q, \tau_{i,<m})}$ is the token-level importance ratio and $\pi_{\text{ref}}$ is the cold-start checkpoint \citep{schulman2017ppo,shao2024deepseekmath,sheng2024verl}.

\subsection{Learning-Frontier Scaling via Structural Tiering and Competence-Frontier Bandit}
\label{sec:curriculum}

The objective in Equation~(\ref{eq:grpo-objective}) normalizes advantages within each question's rollout group. That normalization has a direct mathematical consequence governing how the sampling distribution $\mathcal{P}_{\text{sched}}$ over $\mathcal{Q}_{\text{RL}}$ must be designed when the decision space is enlarged: a question contributes a policy gradient if and only if the $G$ rollouts sampled for it disagree.

\begin{proposition}[Variance Collapse under Group Normalization]
\label{prop:variance}
Let $\hat{A}_i = (r(\tau_i) - \bar{r})/\hat{\sigma}$ be the group-normalized advantage in Equation~(\ref{eq:grpo-objective}) over $G$ rollouts $\{\tau_i\}_{i=1}^G \stackrel{\text{i.i.d.}}{\sim} \pi_\theta(\cdot \mid q)$. Whenever the reward $r(\tau)$ is constant on the sampled support of $\pi_\theta(\cdot \mid q)$ (i.e., $r(\tau_1) = \dots = r(\tau_G) = c$), the within-group variance vanishes ($\hat{\sigma} = 0$) and the policy-gradient contribution of question $q$ is identically zero ($\nabla_\theta \mathcal{J}_q(\theta) \equiv \mathbf{0}$). Furthermore, when format compliance has converged ($\varphi(\tau) = 1.0$) so that $r(\tau) - 1 \sim \text{Bernoulli}(p(q; \theta))$ with success probability $p(q; \theta) \in [0, 1]$, the probability that question $q$ yields a non-zero gradient update across $G$ rollouts is
\begin{equation}
\Pr\big(\hat{\sigma}(q) > 0 \mid \theta\big) = 1 - p(q; \theta)^G - \big(1 - p(q; \theta)\big)^G,
\label{eq:nonzero-var-prob}
\end{equation}
which vanishes at both extremes $p(q; \theta) \in \{0, 1\}$, attains its unique global maximum at the competence frontier $p^*(q; \theta) = 0.5$, and is strictly monotonically increasing in the Bernoulli standard deviation $\sigma_{\text{Bern}}(q; \theta) = \sqrt{p(q; \theta)(1 - p(q; \theta))}$.
\end{proposition}
The proof is in Appendix~\ref{app:proof}. Proposition~\ref{prop:variance} explains why $\meas{64\%}$ of our full-graph training pool yields zero gradient under uniform sampling at $G=8$ (following $p^G + (1-p)^G$ scaling from an empirical $\meas{73\%}$ zero-variance rate observed at $G=4$): mastered questions ($p_q \to 1$) and questions that are unreachable or currently beyond the policy's horizon ($p_q \to 0$) both collapse to $\hat{\sigma}=0$. Because $\hat{\sigma}(q)$ is a per-question property, a batch mixing half $p_q=1$ and half $p_q=0$ questions still produces zero gradient despite a mean pass rate of $0.5$. Sustaining active gradients therefore requires aligning both the \emph{macro progression direction} and the \emph{micro batch composition} $\mathcal{P}_{\text{sched}}$ with the policy's evolving competence frontier ($p_q \approx 0.5$).

\textbf{Model-Independent Structural Tiers and Macro Frontier Direction.} To partition $\mathcal{Q}_{\text{RL}}$ without circular model-based perplexity, we assign each question a deterministic structural tier $d(q) \in \{1, 2, 3, 4\}$ parsed directly from its gold SPARQL topology and official structural annotations: Tier 1 comprises single-hop relation lookups and simple chains, Tier 2 comprises multi-hop \texttt{composition} chains, Tier 3 comprises multi-constraint \texttt{conjunction} intersections, and Tier 4 comprises \texttt{comparative} and \texttt{superlative} aggregation queries. At the start of RL, the policy's competence frontier ($p_q \approx 0.5$) resides in simpler lookups and short chains while harder multi-constraint tiers sit near $p_q \approx 0$. Presenting tiers in progressive order ($d = 1 \to 2 \to 3 \to 4$, \emph{Easy-to-Hard}, E2H) provides an open-loop approximation to the expanding frontier, whereas reversing the exact same $28{,}215$-question multiset ($d = 4 \to 3 \to 2 \to 1$, \emph{Hard-to-Easy}, H2E) front-loads zero-variance hard tiers when the policy cannot solve them and defers easy tiers until they are already mastered.

\textbf{Micro Frontier Adaptation via Self-Evolving Curriculum (\secbandit{}) Bandit.} While static E2H ordering captures the macro directional progression of multi-hop learning, fixed step boundaries cannot adapt to the exact rate at which a policy masters each tier or retreat when the hardest tier remains largely unsolvable. Following \citet{chen2025sec}, we treat the four structural tiers $k \in \{1, 2, 3, 4\}$ over the same $\mathcal{Q}_{\text{RL}}$ pool as bandit arms. Let $\mathcal{B}_{s,k}$ denote the subset of questions drawn from arm $k$ at step $s$, and let $p_q = \frac{1}{G}\sum_{i=1}^G \mathbb{I}[g(\tau_i)\cdot \mathbb{I}[\normset{\hat{A}(\tau_i)} = \normset{A^*(q)}] = 1]$ be the grounded group pass rate of question $q$. Note that while GRPO computes advantages on total reward $r(\tau)$, the scheduler tracks variance strictly on the grounded outcome $p_q$ so that bandit allocation is never misled by formatting noise. \secbandit{} updates each arm's EMA frontier value $Q_k(s)$ and samples arms via a Boltzmann distribution:
\begin{equation}
\begin{gathered}
Q_k(s) = (1 - \alpha) Q_k(s-1) + \alpha \bar{v}_k(s), \qquad \bar{v}_k(s) = \frac{1}{\lvert\mathcal{B}_{s,k}\rvert} \sum_{q \in \mathcal{B}_{s,k}} 2\sqrt{p_q(1-p_q)}, \\
P_{s+1}(k) = \frac{\exp(Q_k(s)/\eta)}{\sum_{j=1}^{4}\exp(Q_j(s)/\eta)},
\end{gathered}
\label{eq:sec-bandit}
\end{equation}
with EMA rate $\alpha = 0.5$, temperature $\eta = 0.4$, and initialization $Q_k(0) = 0.5$ (questions within each arm are drawn without replacement per pass and reshuffled upon depletion). Because $2\sqrt{p_q(1-p_q)}$ decays as an easier tier is mastered ($p_q \to 1$) and rises as a harder tier enters the solvable band ($p_q \to 0.5$), $P_s(k)$ autonomously tracks the moving competence frontier (Algorithm~\ref{alg:training}, Appendix~\ref{app:proof}).

\section{Experiments}
\label{sec:exp}

To systematically evaluate \rise{} and dissect the contribution of each component in its training recipe, we conduct comprehensive experiments across nineteen benchmarks spanning three capability families to answer four research questions (RQs): (1) \textbf{RQ1}: Can effective multi-hop search policies be learned directly over an unpruned, full-scale structured environment, and how well do they generalize to unseen schemas and across model scales? (2) \textbf{RQ2}: Can a search policy trained exclusively on a structured knowledge graph transfer zero-shot to unstructured open-domain passage retrieval? (3) \textbf{RQ3}: Do the hypothesis-testing, pruning, and verification habits acquired through interactive structured exploration generalize out-of-domain to enhance tool-free mathematical and logical reasoning? (4) \textbf{RQ4}: How do the protocol-preserving cold start, progressive frontier ordering, and online variance-driven batch composition (\secbandit{}) distinctively contribute to agentic learning?

\subsection{Experimental Setup}
\label{sec:setup}

\subsubsection{Datasets and Metrics}
We evaluate on nineteen benchmarks under greedy decoding ($T=0$), grouped into three capability families: (1) \emph{structured knowledge} (6 sets)---WebQSP \citep{yih2016webqsp} and CWQ \citep{talmor2018cwq} in-domain, GrailQA \citep{gu2021grailqa}, SimpleQuestions \citep{bordes2015simplequestions}, and WebQuestions \citep{berant2013webquestions} for unseen-schema transfer, and MultiTQ \citep{chen2024multitq} for temporal reasoning; (2) \emph{cross-tool passage retrieval} (7 sets)---the single-fact sets NQ \citep{kwiatkowski2019nq}, TriviaQA \citep{joshi2017triviaqa}, and PopQA \citep{mallen2023popqa}, and the multi-hop sets HotpotQA \citep{yang2018hotpotqa}, 2WikiMultiHopQA \citep{ho2020wiki}, MuSiQue \citep{trivedi2022musique}, and Bamboogle \citep{press2022bamboogle}; and (3) \emph{tool-free general reasoning} (6 sets)---GSM8K \citep{cobbe2021gsm8k}, MATH500 \citep{hendrycks2021math}, AIME24, AMC23, and Knights and Knaves Easy and Hard \citep{xie2024memorization,xie2025logicrl}. Metrics follow standard conventions per family and are never pooled across families: set-equality EM for WebQSP, CWQ, and GrailQA; any-item hit ($\dagger$) for SimpleQuestions and WebQuestions; Hits@1 for MultiTQ; alias EM for passage retrieval; and exact accuracy for general reasoning, with \emph{Avg.} reporting the within-table mean.

\subsubsection{Baselines}
We instantiate every arm on three instruction-tuned backbones---Qwen2.5-7B-Instruct, Qwen2.5-14B-Instruct, and Llama-3.1-8B-Instruct---and compare each capability family against strong specialized baselines reproduced on the exact same backbones and evaluated under identical greedy decoding, prompts (Appendix~\ref{app:prompts}), and answer normalization: Plan-on-Graph (PoG) \citep{chen2024pog} and KG-R1 \citep{lin2025kgr1} for structured knowledge; Search-R1 \citep{jin2025searchr1} and ZeroSearch \citep{sun2025zerosearch} for passage retrieval; and GraphWalker \citep{xu2026graphwalker} and SIE \citep{yu2026sie} for general reasoning. One deliberate asymmetry should be noted: in Tables~\ref{tab:searchqa} and~\ref{tab:reasoning}, \rise{} is evaluated \emph{strictly zero-shot} (trained solely on Freebase), whereas Search-R1 and ZeroSearch are trained in-domain on passage retrieval, making any margin there a conservative lower bound on transfer efficacy. For RQ4 (Table~\ref{tab:arms}), we compare seven controlled recipe arms within the exact same environment: Base model, \rise{} w/o RL (cold start only on $\mathcal{Q}_{\text{cold}}$), \rise{} w/o SFT (RL from base on $\mathcal{Q}_{\text{RL}}$), and full two-stage \rise{} where all four curriculum schedules---uniform (UNI), hard-to-easy (H2E), easy-to-hard (E2H), and \secbandit{} bandit---train on the exact same $28{,}215$-question pool $\mathcal{Q}_{\text{RL}}$.

\subsubsection{Implementation Details}
All RL arms use on-policy multi-turn GRPO on VeRL \citep{sheng2024verl} with vLLM rollouts, group size $G=8$, batch size $64$ questions per step, constant learning rate $1\times10^{-6}$, KL penalty $\beta=0.01$, per-turn cap $2{,}048$ tokens, rolling context window $28{,}672$ tokens, and call budget $C_{\max}=16$ for one epoch ($440$ steps, matching $28{,}215 / 64 \approx 440$) on $8\times\text{A100}$ GPUs. Passage retrieval uses the Wiki-18 dump ($21{,}015{,}324$ passages) indexed with \texttt{E5-base-v2} and FAISS flat, returning top-$3$ passages per call, while the KG adapter returns at most $50$ relations and $100$ rows (Appendix~\ref{app:config}).

\subsection{Main Results}
\label{sec:main-results}

\begin{table}[t]
\centering
\caption{Structured-knowledge results (RQ1), \%. Set-equality EM except SimpleQuestions/WebQuestions (any-item hit, $\dagger$) and MultiTQ (Hits@1). Within each block, \textbf{bold} marks the best result and \underline{underline} the runner-up; \rise{} is shaded. \textbf{\rise{} wins 17 of 18 comparisons, with the largest margins on compositional (CWQ) and unseen-schema (GrailQA) sets.}}
\label{tab:kgqa}
\tabbox{\begin{tabular}{@{}lccccccc@{}}
\toprule
& \multicolumn{2}{c}{In-domain} & \multicolumn{3}{c}{Cross-graph} & Temporal & \\
\cmidrule(lr){2-3}\cmidrule(lr){4-6}\cmidrule(lr){7-7}
Backbone / Arm & WebQSP & CWQ & GrailQA & SimpleQ$^\dagger$ & WebQ$^\dagger$ & MultiTQ & Avg. \\
\midrule
\multicolumn{8}{@{}l}{\emph{Qwen2.5-7B-Instruct}} \\
  Base model & 42.3 & 29.4 & 46.2 & 61.4 & 52.3 & 27.3 & 43.2 \\
  \quad + PoG & 47.5 & 32.1 & 54.3 & 64.2 & 52.7 & 30.2 & 46.8 \\
  \quad + KG-R1 & \second{82.4} & \second{65.6} & \second{62.4} & \second{68.8} & \second{64.5} & \second{33.0} & \second{62.8} \\
  \riserow \quad + \rise{} (ours) & \best{92.4} & \best{85.4} & \best{86.2} & \best{72.2} & \best{67.1} & \best{34.5} & \best{73.0} \\
\midrule
\multicolumn{8}{@{}l}{\emph{Qwen2.5-14B-Instruct}} \\
  Base model & 50.2 & 54.4 & 68.5 & 66.2 & 58.3 & 36.4 & 55.7 \\
  \quad + PoG & 63.4 & 67.6 & 72.3 & 67.4 & 58.9 & 37.2 & 61.1 \\
  \quad + KG-R1 & \second{88.3} & \second{78.4} & \second{79.2} & \second{74.5} & \second{70.0} & \second{42.5} & \second{72.1} \\
  \riserow \quad + \rise{} (ours) & \best{94.6} & \best{89.5} & \best{88.3} & \best{80.8} & \best{71.2} & \best{45.3} & \best{78.3} \\
\midrule
\multicolumn{8}{@{}l}{\emph{Llama-3.1-8B-Instruct}} \\
  Base model & 46.3 & 32.5 & 45.6 & 63.3 & 51.8 & 26.8 & 44.4 \\
  \quad + PoG & 56.9 & 56.8 & 51.3 & 65.8 & 54.4 & 29.1 & 52.4 \\
  \quad + KG-R1 & \second{86.2} & \second{76.1} & \second{77.9} & \second{74.2} & \best{68.4} & \second{32.8} & \second{69.3} \\
  \riserow \quad + \rise{} (ours) & \best{93.2} & \best{88.9} & \best{85.4} & \best{74.3} & \second{68.2} & \best{34.0} & \best{74.0} \\
\bottomrule
\end{tabular}}

\vspace{6pt}

\caption{Zero-shot cross-tool transfer to Wikipedia passage retrieval (RQ2), \% (alias EM). \textbf{\rise{} is evaluated strictly zero-shot (trained only on Freebase); Search-R1 and ZeroSearch are trained in-domain on passage retrieval.} \textbf{\rise{} surpasses domain-trained retrieval baselines on all three backbones, with the largest margins on multi-hop synthesis.}}
\label{tab:searchqa}
\tabbox{\begin{tabular}{@{}lcccccccc@{}}
\toprule
& \multicolumn{3}{c}{Single-fact} & \multicolumn{4}{c}{Multi-hop} & \\
\cmidrule(lr){2-4}\cmidrule(lr){5-8}
Backbone / Arm & NQ & TriviaQA & PopQA & HotpotQA & 2Wiki & MuSiQue & Bamboogle & Avg. \\
\midrule
\multicolumn{9}{@{}l}{\emph{Qwen2.5-7B-Instruct}} \\
  Base model & 12.8 & 35.6 & 13.8 & 16.2 & 22.6 & 6.6 & 24.0 & 18.8 \\
  \quad + Search-R1 & 42.4 & 63.4 & \best{51.6} & 32.8 & 33.2 & 17.4 & 26.4 & 38.2 \\
  \quad + ZeroSearch & \second{43.6} & \best{65.2} & 48.8 & \second{34.6} & \second{35.2} & \second{18.4} & \second{38.4} & \second{40.6} \\
  \riserow \quad + \rise{} (ours) & \best{43.9} & \second{63.8} & \second{49.7} & \best{39.4} & \best{39.3} & \best{18.9} & \best{45.6} & \best{42.9} \\
\midrule
\multicolumn{9}{@{}l}{\emph{Qwen2.5-14B-Instruct}} \\
  Base model & 23.7 & 52.9 & 33.4 & 28.1 & 32.2 & 18.4 & 36.0 & 32.1 \\
  \quad + Search-R1 & 48.8 & 68.7 & 55.2 & 39.2 & 47.9 & 31.4 & 46.4 & 48.2 \\
  \quad + ZeroSearch & \second{58.3} & \second{74.3} & \second{62.9} & \second{43.1} & \second{48.3} & \best{33.6} & \second{50.4} & \second{53.0} \\
  \riserow \quad + \rise{} (ours) & \best{65.4} & \best{76.3} & \best{64.6} & \best{54.3} & \best{57.3} & \second{32.5} & \best{62.4} & \best{59.0} \\
\midrule
\multicolumn{9}{@{}l}{\emph{Llama-3.1-8B-Instruct}} \\
  Base model & 18.5 & 36.6 & 24.3 & 15.4 & 24.9 & 15.4 & 26.4 & 23.1 \\
  \quad + Search-R1 & 41.6 & 62.5 & 50.2 & 42.9 & 39.8 & 23.7 & 40.0 & 42.9 \\
  \quad + ZeroSearch & \second{46.3} & \best{63.0} & \second{52.8} & \second{45.3} & \second{40.5} & \best{24.9} & \second{45.6} & \second{45.5} \\
  \riserow \quad + \rise{} (ours) & \best{50.3} & \second{62.7} & \best{53.6} & \best{48.6} & \best{44.8} & \second{24.6} & \best{51.2} & \best{48.0} \\
\bottomrule
\end{tabular}}
\vspace{-6pt}
\end{table}

\textbf{\rise{} substantially outperforms subgraph-trained and prompting-based KG agents across all three backbones, winning 17 of 18 benchmark comparisons with the largest gains on multi-hop composition and unseen schemas (RQ1).} Table~\ref{tab:kgqa} reports results on the structured-knowledge family. On Qwen2.5-7B, \rise{} elevates the six-benchmark average from $43.2\%$ (Base) to $73.0\%$, outperforming the subgraph-RL baseline KG-R1 ($62.8\%$) by $+10.2$ points and PoG ($46.8\%$) by $+26.2$ points. The advantage is largest precisely where genuine path pruning is required: $92.4\%$ vs.\ $82.4\%$ on WebQSP ($+10.0$), $85.4\%$ vs.\ $65.6\%$ on CWQ ($+19.8$), and $86.2\%$ vs.\ $62.4\%$ on GrailQA ($+23.8$). The same superiority holds on Qwen2.5-14B ($78.3\%$ vs.\ $72.1\%$ for KG-R1) and Llama-3.1-8B ($74.0\%$ vs.\ $69.3\%$). Across datasets, multi-hop compositional queries (CWQ) and zero-shot unseen schemas (GrailQA) exhibit far larger gains than single-hop lookup (SimpleQuestions), confirming that \rise{} learns transferable multi-step pruning rather than superficial query formatting.

\textbf{Policies trained exclusively on Freebase transfer zero-shot to Wikipedia passage retrieval, winning 16 of 21 comparisons and surpassing domain-trained text-RL baselines in overall average across all three backbones (RQ2).} Table~\ref{tab:searchqa} evaluates zero-shot cross-tool transfer when replacing the KG adapter at test time with the Wikipedia dense passage retriever. Without any modification to the policy weights or turn protocol, \rise{} achieves a seven-benchmark average of $42.9\%$ on Qwen2.5-7B (outperforming in-domain trained ZeroSearch at $40.6\%$ and Search-R1 at $38.2\%$), $59.0\%$ on Qwen2.5-14B (vs.\ $53.0\%$ for ZeroSearch), and $48.0\%$ on Llama-3.1-8B (vs.\ $45.5\%$). While multi-hop retrieval benchmarks exhibit the largest gains---lifting HotpotQA from $43.1\%$ to $54.3\%$ ($+11.2$) and Bamboogle from $50.4\%$ to $62.4\%$ ($+12.0$) at 14B---the single-fact benchmarks (NQ, TriviaQA, PopQA) also improve substantially over the base model ($18.8\% \to 42.9\%$ average at 7B), confirming that what transfers across tools is the fundamental habit of query reformulation and evidence verification.

\begin{table}[t]
\centering
\caption{Zero-shot transfer to tool-free general reasoning (RQ3), \% (exact accuracy; AIME24 has $30$ items and AMC23 has $40$, where ties count as neither a win nor a loss). All arms are trained on structured KG environments and evaluated zero-shot without tools. \textbf{Interactive exploration in \rise{} internalizes hypothesis testing and branch pruning, outperforming static in-context RL (SIE) across all three backbones.}}
\label{tab:reasoning}
\tabbox{\begin{tabular}{@{}lccccccc@{}}
\toprule
& \multicolumn{2}{c}{Mathematical} & \multicolumn{2}{c}{Competition} & \multicolumn{2}{c}{Knights \& Knaves} & \\
\cmidrule(lr){2-3}\cmidrule(lr){4-5}\cmidrule(lr){6-7}
Backbone / Arm & GSM8K & MATH500 & AIME24 & AMC23 & Easy & Hard & Avg. \\
\midrule
\multicolumn{8}{@{}l}{\emph{Qwen2.5-7B-Instruct}} \\
  Base model & 30.2 & 42.6 & 3.3 & 37.5 & 41.5 & 18.5 & 28.9 \\
  \quad + GraphWalker & 74.3 & 61.4 & 6.7 & 40.0 & 40.0 & 23.5 & 41.0 \\
  \quad + SIE & \second{87.4} & \second{61.6} & \second{10.0} & \second{47.5} & \best{49.5} & \second{29.0} & \second{47.5} \\
  \riserow \quad + \rise{} (ours) & \best{88.2} & \best{73.4} & \best{16.7} & \best{52.5} & \second{48.5} & \best{31.5} & \best{51.8} \\
\midrule
\multicolumn{8}{@{}l}{\emph{Qwen2.5-14B-Instruct}} \\
  Base model & 71.6 & 62.0 & 16.7 & 57.5 & 60.5 & 34.5 & 50.5 \\
  \quad + GraphWalker & 74.8 & 72.8 & 16.7 & 57.5 & 64.0 & \second{46.5} & 55.4 \\
  \quad + SIE & \best{91.1} & \second{75.0} & \second{20.0} & \second{65.0} & \second{66.0} & 45.5 & \second{60.4} \\
  \riserow \quad + \rise{} (ours) & \second{90.4} & \best{78.8} & \best{30.0} & \best{67.5} & \best{79.0} & \best{65.0} & \best{68.5} \\
\midrule
\multicolumn{8}{@{}l}{\emph{Llama-3.1-8B-Instruct}} \\
  Base model & 66.8 & 38.8 & \second{6.7} & 15.0 & 20.0 & 7.0 & 25.7 \\
  \quad + GraphWalker & 79.2 & \best{48.8} & \second{6.7} & \second{17.5} & 34.0 & 14.0 & 33.4 \\
  \quad + SIE & \second{82.6} & 47.0 & \second{6.7} & \best{22.5} & \second{37.0} & \best{15.5} & \second{35.2} \\
  \riserow \quad + \rise{} (ours) & \best{87.8} & \second{48.4} & \best{10.0} & \best{22.5} & \best{38.5} & \second{15.0} & \best{37.0} \\
\bottomrule
\end{tabular}}

\vspace{6pt}

\caption{Recipe attribution on Qwen2.5-7B-Instruct (RQ4), \%. Six representative benchmarks spanning the three families (this six-column mean is not comparable to the per-family averages in Appendix~\ref{app:sets}). \textbf{Cold start and on-policy RL are complementary, progressive frontier alignment (E2H vs.\ H2E) prevents early zero-variance starvation, and closed-loop \secbandit{} achieves the best overall score.}}
\label{tab:arms}
\tabbox{\begin{tabular}{@{}lccccccc@{}}
\toprule
& \multicolumn{2}{c}{Structured knowledge} & \multicolumn{2}{c}{Retrieval} & \multicolumn{2}{c}{Reasoning} & \\
\cmidrule(lr){2-3}\cmidrule(lr){4-5}\cmidrule(lr){6-7}
Arm & WebQSP & CWQ & PopQA & 2Wiki & GSM8K & MATH500 & Avg. \\
\midrule
  Base model & 42.3 & 29.4 & 13.8 & 22.6 & 30.2 & 42.6 & 30.1 \\
  \quad \rise{} w/o RL & 59.6 & 57.1 & 37.3 & 36.2 & 58.5 & 49.8 & 49.7 \\
  \quad \rise{} w/o SFT & 78.4 & 63.3 & 42.4 & 35.5 & 78.2 & 58.4 & 59.4 \\
  \quad \rise{} w/ UNI & 84.1 & 79.4 & 47.6 & 37.6 & 84.4 & 65.6 & 66.4 \\
  \quad \rise{} w/ H2E & 82.2 & 75.0 & 46.8 & 36.2 & 85.3 & 64.0 & 64.9 \\
  \quad \rise{} w/ E2H & \second{90.7} & \second{83.2} & \best{51.8} & \best{39.8} & \second{87.9} & \second{70.6} & \second{70.7} \\
  \riserow \quad \rise{} w/ \secbandit{} & \best{92.4} & \best{85.4} & \second{49.7} & \second{39.3} & \best{88.2} & \best{73.4} & \best{71.4} \\
\bottomrule
\end{tabular}}
\vspace{-6pt}
\end{table}

\textbf{Interactive structured exploration substantially enhances out-of-domain mathematical and logical reasoning, winning 13 of 18 comparisons (with 1 tie) and surpassing both static in-context RL (SIE) and subgraph RL (GraphWalker) across all three backbones (RQ3).} Table~\ref{tab:reasoning} evaluates zero-shot transfer to six tool-free mathematical and logical reasoning benchmarks. Training in \rise{} elevates the six-benchmark reasoning average from $28.9\%$ to $51.8\%$ on Qwen2.5-7B (surpassing SIE at $47.5\%$ and GraphWalker at $41.0\%$), from $50.5\%$ to $68.5\%$ on Qwen2.5-14B (SIE: $60.4\%$), and from $25.7\%$ to $37.0\%$ on Llama-3.1-8B (SIE: $35.2\%$). The gains are particularly pronounced on multi-step symbolic and logical deduction: on Qwen2.5-7B, \rise{} boosts MATH500 accuracy to $73.4\%$ ($+11.8$ points over SIE), while on Qwen2.5-14B it lifts Knights \& Knaves Easy and Hard to $79.0\%$ and $65.0\%$ ($+13.0$ and $+19.5$ points over SIE, respectively). Because full-graph interaction requires the policy to formulate explicit hypotheses inside \texttt{<think>}, prune invalid branches upon receiving empty or truncated receipts, and verify constraints before answering, these deliberative habits internalize as stronger step-by-step deduction even without external tools.

\textbf{Protocol cold start and on-policy RL are strongly complementary, while progressive frontier alignment (E2H) and closed-loop variance tracking (\secbandit{}) jointly overcome zero-variance stagnation (RQ4).} Table~\ref{tab:arms} ablates each stage and scheduling mechanism across six representative benchmarks on Qwen2.5-7B (full nineteen-benchmark grids are in Appendix~\ref{app:sets}, Tables~\ref{tab:abl-kgqa}--\ref{tab:abl-reasoning}). First, the two stages are mutually reinforcing: supervised cold start alone (\rise{} w/o RL) raises the average from $30.1\%$ to $49.7\%$ and direct RL from the base model (\rise{} w/o SFT) reaches $59.4\%$, whereas combining them under uniform sampling (\rise{} w/ UNI) jumps to $66.4\%$ ($+16.7$ over SFT alone and $+7.0$ over RL alone). Notably, while \rise{} w/o SFT reaches $78.4\%$ on in-domain WebQSP---proving that full-graph tool use is learnable from scratch under our grounded reward---it drops to $59.2\%$ on unseen-schema GrailQA (Appendix Table~\ref{tab:abl-kgqa}), $27.0$ points behind full \rise{} ($86.2\%$), confirming that the cold-start prior is critical for anchoring multi-turn protocol generalization. Second, comparing the curriculum arms over the exact same $28{,}215$-question pool $\mathcal{Q}_{\text{RL}}$ directly validates Proposition~\ref{prop:variance}. Because Tier 3 and Tier 4 questions exhibit near-zero pass rates ($p_q \to 0, \hat{\sigma}=0$) at the start of RL, presenting hard tiers first (\rise{} w/ H2E, $64.9\%$) starves early training of active gradients and underperforms even uniform sampling ($66.4\%$). Conversely, presenting tiers in progressive order (\rise{} w/ E2H, $70.7\%$) acts as an open-loop approximation to the expanding competence frontier, yielding $+4.3$ points over Uniform and $+5.8$ over H2E. Third, replacing fixed tier transitions with closed-loop variance feedback (\rise{} w/ \secbandit{}, $71.4\%$) further refines learning on compositionally demanding benchmarks, advancing CWQ from $83.2\%$ to $85.4\%$ ($+2.2$), WebQSP from $90.7\%$ to $92.4\%$ ($+1.7$), GrailQA from $84.8\%$ to $86.2\%$ ($+1.4$, Table~\ref{tab:abl-kgqa}), MATH500 from $70.6\%$ to $73.4\%$ ($+2.8$), and AMC23 from $47.5\%$ to $52.5\%$ ($+5.0$, Table~\ref{tab:abl-reasoning}).

\subsection{Analysis}
\label{sec:analysis}

\begin{figure}[t]
\centering
\includegraphics[width=0.95\linewidth]{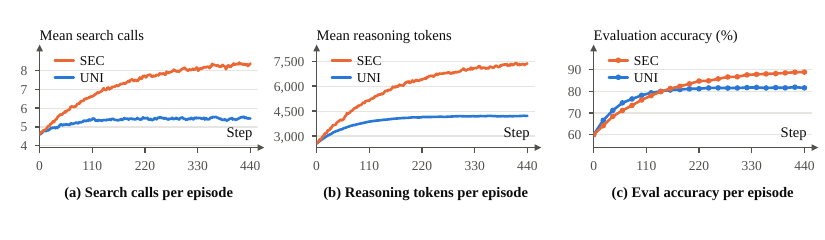}
\vspace{-4pt}
\caption{Training dynamics and sample efficiency comparing competence-frontier scheduling (\secbandit{}) against uniform sampling (UNI) across one epoch ($440$ steps): \textbf{(a)} mean search calls per episode; \textbf{(b)} mean cumulative reasoning tokens per episode; and \textbf{(c)} validation accuracy (\%) on the held-out pool.}
\label{fig:training-dynamics}
\vspace{-6pt}
\end{figure}

\textbf{Frontier-aligned scheduling concentrates its gains on multi-step compositional and unseen-schema bottlenecks.} Inspecting the full nineteen-benchmark ablation grids in Appendix~\ref{app:sets} (Tables~\ref{tab:abl-kgqa}--\ref{tab:abl-reasoning}) reveals where Learning-Frontier Scaling (\rise{} w/ \secbandit{}) extracts its advantage over uniform sampling (\rise{} w/ UNI). On single-hop lookup tasks where uniform sampling already encounters frequent non-zero variance early in training, the margin between \secbandit{} and UNI is modest ($+1.8$ points on SimpleQuestions and $+1.5$ on WebQuestions). In contrast, on tasks requiring multi-hop relational composition, zero-shot schema generalization, or deep symbolic chains---where uniform rollouts frequently collapse to all-zero rewards early on---the margin expands dramatically: $+8.3$ points on WebQSP ($92.4\%$ vs.\ $84.1\%$), $+6.0$ on CWQ ($85.4\%$ vs.\ $79.4\%$), $+21.6$ on GrailQA ($86.2\%$ vs.\ $64.6\%$), $+7.8$ on MATH500 ($73.4\%$ vs.\ $65.6\%$), and $+7.5$ on AMC23 ($52.5\%$ vs.\ $45.0\%$).

\textbf{Co-evolution of search depth, episode deliberation, and sample efficiency.} Figure~\ref{fig:training-dynamics} tracks the training dynamics of \secbandit{} versus uniform sampling (UNI) across the $440$ optimizer steps of one epoch (see Appendix~\ref{app:case} for a qualitative trajectory comparison). In Figure~\ref{fig:training-dynamics}a, both schedules begin at an average of $4.6$ search calls per episode; however, while UNI plateaus early around $5.4$ calls, \secbandit{} progressively expands its interaction horizon to $8.3$ search calls per episode as bandit mass migrates from single-hop lookups (Tier 1) toward multi-hop chains and conjunctive intersections (Tiers 2--3). Correspondingly, Figure~\ref{fig:training-dynamics}b shows that cumulative reasoning tokens per episode grow from $\sim 2{,}600$ to $\sim 7{,}400$ under \secbandit{} (versus plateauing at $\sim 4{,}200$ under UNI), confirming that decoupling the call budget ($C_{\max}=16$) from the per-turn token cap ($2{,}048$ tokens) enables deeper multi-turn verification without per-turn truncation. Finally, Figure~\ref{fig:training-dynamics}c plots validation accuracy on the held-out WebQSP + CWQ test pool ($5{,}156$ questions): although UNI rises slightly faster in the first $100$ steps by harvesting easy lookups, it saturates near $81.8\%$ by step $180$. By contrast, \secbandit{} overtakes UNI at step $\sim 145$, matches UNI's full-epoch accuracy within $38\%$ of the training budget ($\sim 165$ steps), and continues climbing to $\sim 89.0\%$ across the epoch.

\section{Conclusion}
\label{sec:conclusion}

What defines an autonomous retrieval agent is the sequence of decisions it makes: which path to explore, which candidates to prune, and when to stop searching. All three decisions vanish when the training environment pre-extracts a local subgraph or lets an upstream index decide what is reachable in advance. \rise{} transitions structured RL environments to full-scale interactive databases along two coupled axes: \emph{Decision-Space Scaling} exposes the unpruned knowledge graph through a five-stage graph-agnostic typed intent adapter under a bounded call budget, restoring pruning and stopping as turn-attributable policy decisions; and \emph{Learning-Frontier Scaling} overcomes the $64\%$ within-group variance collapse of full-graph exploration by aligning progressive tier ordering and closed-loop \secbandit{} batch composition with the policy's evolving competence frontier. Across nineteen benchmarks and three model backbones, \rise{} substantially outperforms specialized baselines in structured KG reasoning, transfers zero-shot to Wikipedia passage retrieval, and generalizes out-of-domain to enhance tool-free mathematical and logical deduction.

\newpage

\subsubsection*{Ethics Statement}
This work trains and evaluates retrieval-augmented language models exclusively over publicly released benchmarks and knowledge bases (Freebase, MultiTQ, and Wikipedia), involving no human subjects or private user data. While our multiplicative grounding gate $g(\tau)$ substantially suppresses ungrounded fabrication, a grounded policy can still misinterpret retrieved facts; we report unconditional full-split metrics and cross-family evaluations transparently to support safe deployment.

\subsubsection*{Reproducibility Statement}
To ensure full reproducibility, Appendix~\ref{app:related} reviews related work, Appendix~\ref{app:prelim} details our preliminary subgraph measurements and design requirements, Appendix~\ref{app:contract} specifies the formal environment contract and closed error vocabulary, Appendix~\ref{app:prompts} reproduces verbatim prompt templates for all five task families, Appendix~\ref{app:config} lists all training, data-split, and retrieval hyperparameters, Appendix~\ref{app:sets} reports the complete nineteen-benchmark attribution grids, Appendix~\ref{app:case} provides a side-by-side multi-turn interaction case study, Appendix~\ref{app:proof} gives the proof of Proposition~\ref{prop:variance} and Algorithm~\ref{alg:training}, Appendix~\ref{app:discussion} and Appendix~\ref{app:limitations} discuss broader methodological principles and limitations, and Appendix~\ref{app:defects} documents our 23-item silent-failure audit taxonomy. Source code and environment adapters will be released upon publication.

\bibliography{refs}
\bibliographystyle{iclr2027_conference}

\newpage
\appendix
\input{appendix}

\end{document}

%% file: math_commands.tex
\usepackage{amsmath,amsfonts,bm}

\def\eqref#1{equation~\ref{#1}}

\def\1{\bm{1}}

\DeclareMathAlphabet{\mathsfit}{\encodingdefault}{\sfdefault}{m}{sl}
\SetMathAlphabet{\mathsfit}{bold}{\encodingdefault}{\sfdefault}{bx}{n}



%% file: appendix.tex
%
%

\addtocontents{toc}{\protect\setcounter{tocdepth}{2}}
\renewcommand{\contentsname}{Appendix Contents}
\tableofcontents
\newpage

\section{Related Work}
\label{app:related}

Table~\ref{tab:feature-matrix} summarizes how representative systems sit relative to \rise{} along the three core environmental properties established in Section~\ref{sec:intro}.

\begin{table}[!htbp]
\centering
\caption{Positioning of representative methods along the three environmental properties argued in this paper. A filled checkmark shows that the property holds by construction; an open circle shows that it holds only partially (e.g., operating on restricted or synthetic subgraphs, or annealing an exogenous noise knob rather than tracking policy variance).}
\label{tab:feature-matrix}
\tabbox{\begin{tabular}{@{}llccc@{}}
\toprule
Method & Action interface & Full-graph interaction & Graph-agnostic intent & Variance-driven curriculum \\
\midrule
Think-on-Graph \citep{sun2024tog} & Beam search over edges & $\checkmark$ & $\circ$ & -- \\
Plan-on-Graph \citep{chen2024pog} & LLM-guided expansion & $\circ$ & $\circ$ & -- \\
KG-R1 \citep{lin2025kgr1} & Four symbolic one-hop ops & -- & $\circ$ & -- \\
GraphWalker \citep{xu2026graphwalker} & Relation / triple lookup & $\circ$ & $\circ$ & -- \\
Search-R1 \citep{jin2025searchr1} & Passage query & -- & $\checkmark$ & -- \\
ZeroSearch \citep{sun2025zerosearch} & Passage query & -- & $\checkmark$ & $\circ$ \\
SIE \citep{yu2026sie} & Static in-context subgraph & -- & -- & -- \\
\midrule
\riserow \rise{} (ours) & Natural-language intent & $\checkmark$ & $\checkmark$ & $\checkmark$ \\
\bottomrule
\end{tabular}}
\end{table}

\paragraph{Knowledge-Graph Reasoning Agents.} Prompt-driven frameworks traverse a graph by prompting a frozen model to select edges iteratively: Think-on-Graph \citep{sun2024tog} performs beam search over relations, and Plan-on-Graph \citep{chen2024pog} maintains a decomposed plan with reflection and adaptive beam width. A second line retrieves a question-specific subgraph prior to generation, as in Reasoning-on-Graphs \citep{luo2024rog} and G-Retriever \citep{he2024gretriever}. A third line trains the policy end to end via RL: KG-R1 \citep{lin2025kgr1} optimizes multi-turn retrieval with four symbolic one-hop primitives over pre-extracted per-question subgraphs; GraphWalker \citep{xu2026graphwalker} pretrains on synthetic walking trajectories and applies RL over relation and triple lookups; KnowCoder-A1 \citep{chen2025knowcoder} emits executable SPARQL queries assisted by schema-lookup tools; ISP-KGR \citep{song2026ispkgr} couples interactive semantic parsing with tree-structured rollouts; and Temp-R1 \citep{gong2026tempr1} applies a reverse curriculum over temporal primitives. Two fundamental differences separate \rise{} from this line. First, regarding \emph{decision-space scale}, a median per-question subgraph in the KG-R1 release holds only $\meas{4{,}415}$ triples, enclosing the gold path in advance; \rise{} exposes the full $\meas{9.5\times10^{8}}$-triple store under a bounded call budget. Notably, \citet{xu2026graphwalker} report that evaluating subgraph-trained KG-R1 on the global graph drops WebQSP accuracy from $82.7\%$ (in-subgraph) to $66.8\%$, directly confirming the magnitude of the subgraph shortcut. Second, regarding \emph{interface abstraction}, SPARQL generation and graph-specific primitives bind the policy to one database schema, whereas our natural-language intent adapter transfers zero-shot to unstructured passage retrieval.

\paragraph{Reinforcement Learning for Search and Tool Use.} Interleaving chain-of-thought reasoning with external tool calls \citep{yao2023react,schick2023toolformer} via outcome-rewarded RL has been widely studied on text corpora: Search-R1 \citep{jin2025searchr1} formulates multi-turn passage retrieval with observation-token masking, while ReSearch \citep{chen2025research}, R1-Searcher \citep{song2025r1searcher}, WebThinker \citep{li2025webthinker}, and DeepResearcher \citep{zheng2025deepresearcher} extend outcome-driven RL to multi-step web and corpus search. ToolRL \citep{qian2025toolrl} studies reward granularity for general API invocation, SimpleTIR \citep{xue2026simpletir} stabilizes multi-turn tool-integrated training by filtering degenerate turns, Explore-on-Graph \citep{yan2026eog} introduces path-refined reward modeling, and ZeroSearch \citep{sun2025zerosearch} replaces live search engines with a learned simulator whose noise is annealed over training. While effective on homogeneous text corpora where every query returns a stateless top-$k$ list with smooth similarity decay, directly applying text-RL recipes to a full-scale structured knowledge base encounters ungrounded guessing on multi-answer queries and severe zero-variance gradient starvation---which \rise{} resolves via our multiplicative grounding gate $g(\tau)$ and competence-frontier scheduling.

\paragraph{Structured Post-Training Environments and Curriculum Learning.} A parallel research line constructs post-training environments from structured data \citep{fang2025general,lacombe2025reasoning,wang2025graphr1}. KaLM \citep{yu2026kalm} aligns internal LLM representations with KG triples via dual-view contrastive learning, embedding structured knowledge into model parameters rather than context. SIE \citep{yu2026sie} automatically extracts static in-context subgraphs of controlled completeness ($\rho \equiv 1$ at SIE-100\%) to incentivize compositional deduction via GRPO. ReMA \citep{wan2025rema} explores multi-agent meta-thinking under RL, \citet{wen2024reinforcing} study action decomposition for language agents, and Logic-RL \citep{xie2025logicrl} demonstrates reasoning transfer from rule-based logic puzzles. The key distinction lies in what is scaled: \emph{SIE scales the completeness of a static context that the model passively reads, whereas \rise{} scales the decision space of an interactive search environment that the model actively queries}, where reachability is governed by the policy's own sequential decisions under a call budget. On the optimization side, classical curriculum learning orders examples statically by difficulty \citep{bengio2009curriculum,kumar2010selfpaced}, whereas prioritized replay and bandit schedulers adapt sampling online \citep{schaul2016prioritized,auer2002ucb}. Building upon \citet{chen2025sec}, we ground curriculum design in Proposition~\ref{prop:variance} by uniting SPARQL structural tiers with the group-level standard deviation return $2\sqrt{p_q(1-p_q)}$.

\section{Preliminary Measurements and Environment Design Requirements}
\label{app:prelim}

This appendix provides the complete empirical protocols and measurements underlying the environment design of \rise{}. Specifically, we formalize three measurable criteria that an interactive environment must satisfy to support authentic search-policy learning, evaluate existing subgraph and query-syntax paradigms against these criteria across four diagnostic measurement sets, and derive the six engineering and methodological requirements summarized in Table~\ref{tab:requirements}.

\subsection{Three Measurable Criteria for Agentic Search Environments}
Treating multi-turn retrieval as a learnable decision process requires an environment that simultaneously satisfies three foundational conditions:

\textbf{(C1) Unavoidable Candidate Pruning.} If the candidate space is small enough to enumerate within a few steps, sequential path selection requires no strategic pruning. Given any candidate environment, we audit C1 by measuring: (a) the fraction of ground-truth answers reachable within one or two hops from the topic entity, and (b) the top-$k$ answer hit rate achievable by a question-blind structural heuristic that never reads the input question. If (b) substantially exceeds uniform random chance, the environment's topology leaks the answer's location.

\textbf{(C2) Graph-Agnostic Action Interface.} If the policy's action space is bound to a database-specific query syntax (e.g., SPARQL or specialized atomic graph signatures), the policy conflates strategic search planning with syntax generation, obstructing zero-shot transfer to new graph schemas or unstructured text retrievers. C2 is satisfied when the policy emits natural-language search intents while a deterministic environment adapter handles schema-specific translation.

\textbf{(C3) Turn- and Outcome-Level Attributability.} If the training reward incorporates intermediate process shaping (e.g., bonus credit whenever a gold entity appears anywhere in a retrieved table), score improvements cannot be cleanly attributed to improved decision-making versus shaping exploitation. Attributability requires an outcome-only reward gated by explicit evidence grounding, paired with closed-vocabulary turn receipts.

\subsection{Measurement 1: Per-Question Subgraphs Compress the Decision Space and Leak Answer Locations}
We audited the per-question WebQSP subgraphs released by KG-R1 \citep{lin2025kgr1} across all test instances without modification. Hop counts are computed via breadth-first search (BFS) as the shortest path length from any linked topic entity to any gold answer entity inside the extracted subgraph. To test for topological leakage, we evaluate a \emph{question-blind degree heuristic} that ranks all nodes in the subgraph strictly by their in-subgraph degree (breaking ties deterministically by entity identifier) without accessing the question text, compared against a uniform random node baseline averaged over $1{,}000$ Monte Carlo trials per question.

The empirical findings reveal three severe confounds:
\begin{itemize}[leftmargin=1.5em,topsep=2pt,itemsep=2pt]
\item \textbf{Five-Order-of-Magnitude Space Compression:} The median per-question subgraph contains only $\meas{4{,}415}$ triples spanning $\meas{1{,}638}$ entities (subject to a hard cap of $2{,}000$ entities), which is $\meas{216{,}202}\times$ smaller than the full Freebase store ($\meas{9.5\times 10^8}$ triples).
\item \textbf{Enumeration Replaces Multi-Hop Planning:} Exactly $\meas{65.5\%}$ of gold answers are reachable within a single hop and $\meas{29.1\%}$ within two hops. Thus, a naive two-hop expansion covers $\meas{94.6\%}$ of all test questions, eliminating the need for selective path pruning.
\item \textbf{Topological Leakage of Answer Entities:} The question-blind degree heuristic ranks a gold answer entity at Top-1 in $\meas{9.41\%}$ of questions, Top-5 in $\meas{24.11\%}$, and Top-10 in $\meas{33.27\%}$, compared to $\meas{0.19\%}$ for uniform random selection. Across the dataset, gold answer entities exhibit a median in-subgraph degree of $\meas{11}$, whereas random subgraph nodes exhibit a median degree of $\meas{3}$.
\end{itemize}
Because upstream subgraph extractors are designed to enclose the ground-truth reasoning path within a small node budget, they inevitably pull in dense local connectivity around the target answer. Consequently, models trained or evaluated on pre-extracted subgraphs can achieve high accuracy by exploiting construction artifacts rather than learning generalizable search policies.

\subsection{Measurement 2: Database-Specific Action Signatures Obstruct Cross-Tool Transfer}
Binding the policy to fixed atomic graph primitives or raw query languages prevents zero-shot generalization across heterogeneous tools. For instance, extending atomic neighborhood actions from static Freebase to temporal knowledge graphs requires altering the action signature to accept timestamp arguments and introducing specialized temporal operators \citep{lin2025kgr1,gong2026tempr1}. Conversely, semantic parsing agents such as KnowCoder-A1 \citep{chen2025knowcoder} train the policy to emit raw SPARQL syntax supported by schema-lookup tools. While effective within a single database schema, neither action space can be transferred zero-shot to an unstructured Wikipedia passage retriever. By requiring the policy to formulate queries as natural-language intents inside \texttt{<search>...</search>} and delegating execution to a deterministic typed intent adapter $\Phi_{\mathcal{G}}$, \rise{} maintains an identical policy-side protocol across static KGs, temporal KGs, and dense passage retrievers.

\subsection{Measurement 3: Full-Graph Latency, Partial Observability, and Directional Asymmetry}
A common objection to full-graph RL training is that executing live queries against $9.5\times 10^8$ triples during on-policy rollouts will bottleneck training throughput. Profiling our Virtuoso backend refutes this concern: on the execution path, a single adapter \texttt{search} call has a median latency of $\meas{15.5}$~ms and a p95 of $\meas{65.4}$~ms (with worst-case single-query latency staying below one second), ensuring that no individual call stalls a rollout turn. At the level of a training step, a warm relation vocabulary of $6{,}102$ entries sustains $\meas{29.22}$ queries per second at a median latency of $\meas{0.985}$~s, meaning that $2{,}560$ queries---the typical load of one GRPO step at $64$ questions $\times$ $8$ rollouts $\times$ $5$ turns---require only $\meas{1.46}$ minutes of cumulative backend execution time across parallel workers while multi-turn LLM decoding and FSDP parameter updates dominate step wall-clock time.

Operating over the full graph surfaces two structural properties that directly shape our adapter and curriculum design:
\begin{enumerate}[leftmargin=1.6em,topsep=2pt,itemsep=2pt]
\item \textbf{Budget-Bounded Reachability Floor and Early Reward Sparsity:} Let $A^*(q)$ be the ground-truth answer set for question $q$, and $\mathcal{E}(o_t)$ the set of surface-form entities and literal values returned in receipt $o_t$. For an action sequence $a_{1:k} = (a_1, \dots, a_k)$, the budget-bounded reachability indicator $\rho(q; C_{\max}) \in \{0,1\}$ is defined as
\begin{equation}
\rho(q; C_{\max}) = \mathbb{I}\!\left[ \exists\, a_{1:k} \in \mathcal{A}_{\text{search}}^k \ (k \le C_{\max}) \ \text{s.t.}\ \normset{A^*(q)} \subseteq \bigcup_{t=1}^{k} \normset{\mathcal{E}\big(\Phi_{\mathcal{G}}(a_t, h_{t-1})\big)} \right]\!.
\label{eq:reachability}
\end{equation}
Unlike pre-extracted subgraphs where $\rho(q; C_{\max}) \equiv 1$ by construction, full-graph search under $C_{\max}=16$ is subject to partial observability: a model-free outgoing BFS probe without semantic ranking establishes a strict lower bound of $\rho_{\text{BFS}} \ge \meas{0.5219}$ (leaving up to $\meas{0.48}$ unreachable without semantic ranking). Under $G=8$ group sampling, this partial observability causes $\meas{64\%}$ of training questions to yield zero within-group reward variance ($\hat{\sigma}=0$, corresponding to $p^G + (1-p)^G$ scaling from an empirical $\meas{73\%}$ zero-variance rate at $G=4$), motivating our \secbandit{} bandit scheduler.
\item \textbf{Severe Directional Asymmetry of Relations:} Auditing candidate Freebase relations around topic entities reveals that $\meas{48.6\%}$ of valid relations exist exclusively as \emph{incoming} edges $(e', r, e_t)$. An environment adapter that traverses only outgoing edges $(e_t, r, e')$ would list these relations during schema inspection but return empty results when the policy queries them, penalizing valid reasoning steps. Direction-symmetric execution (Equation~(\ref{eq:bidirectional-exec})) is therefore a foundational correctness requirement.
\end{enumerate}

\subsection{Measurement 4: Silent Failure Modes in Multi-Turn Agentic RL}
During the development of \rise{}, we conducted systematic turn-level trajectory audits and discovered that the most damaging implementation defects in multi-turn agentic RL do not trigger runtime exceptions or degrade standard scalar monitoring curves (e.g., mean reward, KL divergence, or gradient norm). Instead, defects such as mid-row observation truncation, test-split leakage into cold-start pools, reward-gaming on multi-answer sets, or mismatches between trainer and evaluator stop tokens manifest as \emph{silent capability regressions} where training jobs complete normally (\texttt{Succeeded}) while downstream transfer or reasoning retention degrades. We formalize these into a 23-item audit taxonomy in Appendix~\ref{app:defects}.

\begin{table}[!htbp]
\centering
\caption{Summary of environment and training design requirements derived from Measurements 1--4.}
\label{tab:requirements}
\tabbox{
\begin{tabular}{@{}p{0.05\linewidth}>{\raggedright\arraybackslash}p{0.31\linewidth}>{\raggedright\arraybackslash}p{0.20\linewidth}>{\raggedright\arraybackslash}p{0.36\linewidth}@{}}
\toprule
ID & Design Requirement & Empirical Source & Architectural Instantiation in \rise{} \\
\midrule
D1 & Unpruned candidate space requiring genuine path pruning & Criterion C1, Measurement 1 & Full-scale Freebase ($9.5\times 10^8$ triples); zero per-question subgraph pre-filtering \\
D2 & Graph- and tool-agnostic policy action interface & Criterion C2, Measurement 2 & Natural-language \texttt{<search>} intents translated by deterministic adapter $\Phi_{\mathcal{G}}$ \\
D3 & Outcome-only reward immune to parametric guessing & Criterion C3, Section~\ref{sec:reward-choice} & Set-equality EM gated multiplicatively by evidence grounding $g(\tau)$ + additive format $\varphi(\tau)$ \\
D4 & Deterministic, answer-blind, direction-symmetric execution & Measurement 3 & Bidirectional triple traversal $\mathcal{F}_{\text{exec}}$ with explicit truncation headers \texttt{(k of N)} \\
D5 & Protocol bootstrapping and active gradient maintenance & Measurement 3, Proposition~\ref{prop:variance} & Isomorphic teacher rejection-sampled cold start + SPARQL tiers \& \secbandit{} bandit \\
D6 & Multi-family evaluation without cross-family pooling & Measurement 4, Appendix~\ref{app:defects} & Three capability families across nineteen benchmarks, never pooled across families \\
\bottomrule
\end{tabular}}
\end{table}

\section{Formal Environment Contract and Error Vocabulary}
\label{app:contract}

This appendix specifies the formal turn grammar, receipt schema, closed error vocabulary, and budget accounting rules summarized in Section~\ref{sec:env}.

\paragraph{Turn Grammar.} At every turn $t$, the assistant policy must emit a reasoning block enclosed in \texttt{<think>...</think>} followed by exactly one action tag (either a search query or a terminal answer):
{\small
\begin{Verbatim}
<think>step-by-step reasoning and search planning</think>
<search>natural-language query with [[Entity Mention]]</search>

<think>final evidence synthesis and verification</think>
<answer>entity 1, entity 2</answer>
<answer></answer>   % explicit abstention
\end{Verbatim}
}
Every \texttt{<search>} turn is followed by exactly one \texttt{user} turn containing the environment's \texttt{<information>...</information>} receipt; retrieved facts are never spliced directly into the assistant's generation stream.

\paragraph{Structured Receipt Schema.} Each valid observation receipt contains four standardized fields: (1) \texttt{Resolved}, reporting the surface form and matched graph node identifier (allowing the policy to verify entity resolution without generating raw MIDs); (2) \texttt{intent}, recording the classified structural intent tag ($\iota_t \in \{\text{\texttt{relation\_hop}}, \text{\texttt{relation\_list}}, \text{\texttt{entity\_info}}\}$); (3) \texttt{Results (k of N)}, listing the top-$k$ ranked triples formatted as \texttt{subject | relation | object} alongside total match count $N$ to distinguish empty results ($N=0$) from row-cap truncation ($k < N$); and (4) \texttt{Calls remaining}, displaying the remaining search budget $C_{\max} - c_t$.

\paragraph{Closed-Vocabulary Error Codes.} All execution anomalies are mapped deterministically to seven machine-readable error codes:
\begin{itemize}[leftmargin=1.5em,topsep=2pt,itemsep=1pt]
\item \texttt{E\_NO\_ENTITY}: Query contains no bracketed entity mention or no candidate exceeds the resolution threshold.
\item \texttt{E\_ENTITY\_NOT\_FOUND}: Bracketed surface form cannot be resolved in the active binding table or graph index.
\item \texttt{E\_INTENT\_UNSUPPORTED}: Requested structural operation is inapplicable to the resolved node type.
\item \texttt{E\_NO\_RESULTS}: Well-formed query executed successfully but returned zero matching triples ($N=0$).
\item \texttt{E\_QUERY\_FAILED}: Transient backend timeout or SPARQL engine exception; never penalized or counted against policy error streaks.
\item \texttt{E\_BUDGET\_EXCEEDED}: Search call cap $C_{\max}=16$ reached; triggers the final answer-only turn.
\item \texttt{E\_MALFORMED}: Assistant output violated the turn grammar (\texttt{<think>} followed by \texttt{<search>} or \texttt{<answer>}); permits retry on the next turn if budget remains.
\end{itemize}

\paragraph{Dual-Budget Accounting and Observation Independence.} The search call cap ($C_{\max}=16$), per-turn generation limit ($2{,}048$ tokens), and rolling context window ($28{,}672$ tokens, enforced via left truncation) operate orthogonal to one another. Cold-start trajectory filtering, online GRPO rollouts, and benchmark evaluation invoke the exact same adapter binary and receipt serializer.

\section{Verbatim Prompt Templates}
\label{app:prompts}

To guarantee full reproducibility of our protocol layer and cross-tool transfer experiments, we reproduce verbatim every system, user, and observation prompt template used across the five task instantiations. All templates share an isomorphic turn structure (\texttt{<think>} $\to$ \texttt{<search>}/\texttt{<answer>} $\to$ \texttt{<information>}), differing only in tool specification and answer formatting.

\subsection{Knowledge-Graph Agent (WebQSP, CWQ, GrailQA, SimpleQuestions, WebQuestions)}
When a benchmark provides linked topic entities, they are supplied in double square brackets on the second line of the user turn; when no topic entities are provided, the second line is omitted.
\begin{promptbox}{System \& User Prompt: Knowledge-Graph Agent}
\begin{Verbatim}
You are a question-answering agent with access to a knowledge graph.

In every turn, first reason inside <think>...</think>, then take exactly one action:
- <search>query</search> to query the knowledge graph, or
- <answer>final answer</answer> to finish.
After a search, the environment replies inside <information>...</information>; wait for that reply before acting again.

Write a search as a natural-language question about one entity, with every entity name in double square brackets, copied exactly from the question, the topic entities or an earlier <information> block:
  <search> Who is the head coach of [[Pittsburgh Steelers]]? </search>

A reply starts with a Resolved line naming the matched entity, followed by facts as "subject | relation | object" rows. "Results (k of N)" means only k of N rows are shown; ask a more specific question to narrow them down. If no relation matches your question, the reply lists the entity's relations instead; choose one and ask again. Repeating a search returns the same result. When a reply says your budget is exhausted, give your answer in the next turn.

Answer with names, dates or values exactly as they appear in <information> blocks, and separate multiple answers with commas.
\end{Verbatim}
\tcblower
\begin{Verbatim}
Question: <question, whitespace-normalised>

Topic entities: [[entity 1]], [[entity 2]]
\end{Verbatim}
\end{promptbox}

\subsection{Temporal Knowledge-Graph Modifier (MultiTQ)}
For MultiTQ, the following paragraph is appended to the knowledge-graph system prompt above:
\begin{promptbox}{System Prompt Modifier: Temporal Knowledge-Graph (MultiTQ)}
\begin{Verbatim}
This knowledge graph is temporal: every fact row ends with the date it happened (YYYY-MM-DD). Answer questions about "before", "after", "first" or "last" by comparing those dates, and give a date at the granularity the question asks for: YYYY for a year, YYYY-MM for a month, YYYY-MM-DD for a day.
\end{Verbatim}
\end{promptbox}

\subsection{Text-Retrieval Agent (Search QA Transfer Family)}
For zero-shot evaluation on NQ, TriviaQA, PopQA, HotpotQA, 2WikiMultiHopQA, MuSiQue, and Bamboogle, the KG adapter is replaced by a Wikipedia passage retriever using the following minimally adapted system prompt:
\begin{promptbox}{System Prompt: Text-Retrieval Agent (Search QA Family)}
\begin{Verbatim}
You are a question-answering agent with access to a Wikipedia search engine.

In every turn, first reason inside <think>...</think>, then take exactly one action:
- <search>query</search> to search Wikipedia, or
- <answer>final answer</answer> to finish.
After a search, the environment replies inside <information>...</information>; wait for that reply before acting again.

Write a search as a natural-language question or a keyword query:
  <search> Who is the head coach of the Pittsburgh Steelers? </search>

A reply lists the most relevant passages, each with its title. If they do not contain what you need, search again with a different query; repeating a search returns the same result. When a reply says your budget is exhausted, give your answer in the next turn.

Answer with a short answer only (a name, date or number), not a sentence.
\end{Verbatim}
\end{promptbox}

\subsection{Single-Turn Mathematical Reasoning (GSM8K, MATH500, AIME24, AMC23)}
\begin{promptbox}{System \& User Prompt: Single-Turn Mathematical Reasoning}
\begin{Verbatim}
You are a careful mathematical problem solver.

First reason step by step inside <think>...</think>, then give the final answer inside <answer>...</answer>. The answer must be the final result only (a number or a simplified expression in LaTeX), with no explanation.
\end{Verbatim}
\tcblower
\begin{Verbatim}
Question: <question>
\end{Verbatim}
\end{promptbox}

\subsection{Logical Reasoning (Knights and Knaves Easy \& Hard)}
\begin{promptbox}{System \& User Prompt: Logical Reasoning (Knights and Knaves)}
\begin{Verbatim}
You are a careful logical reasoner.

First reason step by step inside <think>...</think>, then give the final answer inside <answer>...</answer>, stating the identity of every inhabitant on its own numbered line:
<answer>
(1) A is a knight
(2) B is a knave
</answer>
Use the inhabitants' names from the question in place of A and B.
\end{Verbatim}
\tcblower
\begin{Verbatim}
Question: <question>
\end{Verbatim}
\end{promptbox}

\subsection{Environment Observation Receipt Template}
\begin{promptbox}{Observation Receipt Template}
\begin{Verbatim}
<information>
Resolved: <matched entity> | intent=<closed tag>
Results (k of N):
  <subject> | <relation> | <object>
  ...
Calls remaining: <n>
</information>
\end{Verbatim}
\end{promptbox}

\section{Training Configuration and Hyperparameters}
\label{app:config}

All training instances in Stage 1 and Stage 2 are drawn exclusively from the official training splits of WebQSP and CWQ ($34{,}215$ questions in total), which are partitioned via stratified sampling by dataset and difficulty tier into the cold-start pool $\mathcal{Q}_{\text{cold}}$ ($6{,}000$ questions) and the RL pool $\mathcal{Q}_{\text{RL}}$ ($28{,}215$ questions). The four RL curriculum schedules (UNI, E2H, H2E, and \secbandit{}) operate over the exact same $28{,}215$-question pool $\mathcal{Q}_{\text{RL}}$, differing only in presentation order or dynamic batch composition. Table~\ref{tab:config} details the complete training, data-hygiene, and retrieval configuration shared across all experimental arms.

\begin{table}[!htbp]
\centering
\caption{Full training, data-split, adapter, and evaluation hyperparameters for \rise{}.}
\label{tab:config}
\tabbox{
\begin{tabular}{@{}l>{\raggedright\arraybackslash}p{0.62\linewidth}@{}}
\toprule
Hyperparameter / Setting & Value \\
\midrule
Base models & Qwen2.5-7B-Instruct (primary); Llama-3.1-8B-Instruct; Qwen2.5-14B-Instruct \\
RL optimization framework & On-policy multi-turn GRPO on VeRL \citep{sheng2024verl} (FSDP + vLLM rollout engine) \\
Rollout group size $G$ & 8 trajectories per question \\
Batch size $\lvert\mathcal{B}_s\rvert$ & 64 questions per step ($512$ group trajectories per step) \\
Training horizon $S_{\text{epoch}}$ & 1 epoch ($440$ optimizer steps) \\
Master training corpus & $34{,}215$ questions from official WebQSP + CWQ train splits, stratified by \texttt{(dataset, tier)} into strictly disjoint $\mathcal{Q}_{\text{cold}}$ ($6{,}000$) and $\mathcal{Q}_{\text{RL}}$ ($28{,}215$) \\
Cold-start pool $\mathcal{Q}_{\text{cold}}$ \& teacher & $6{,}000$ questions; Gemini-2.5-pro ($T=0.7$, $\le 2$ attempts/question; $\lvert\mathcal{D}_{\text{SFT}}\rvert=4{,}006$ accepted under Eq.~\ref{eq:sft-corpus}) \\
Cold-start SFT epochs & 2 epochs \\
RL question pool $\mathcal{Q}_{\text{RL}}$ & $28{,}215$ questions (shared identically across UNI, E2H, H2E, and \secbandit{}; disjoint from $\mathcal{Q}_{\text{cold}}$) \\
Periodic validation split & Official WebQSP + CWQ test splits ($1{,}628 + 3{,}528 = 5{,}156$ questions, evaluated every $50$ steps) \\
Learning rate & $1\times 10^{-6}$ (constant, AdamW) \\
KL divergence penalty $\beta$ & $0.01$ against $\pi_{\text{ref}}$ \\
Clip ratio $\epsilon$ / Entropy bonus & $\epsilon = 0.2$ / Entropy bonus $= 0.0$ \\
Per-turn generation cap & $2{,}048$ tokens \\
Rolling context window & $28{,}672$ tokens (left truncation) \\
Search call budget $C_{\max}$ & $16$ calls per episode \\
KG adapter ranking \& caps & Dense encoder \texttt{intfloat/e5-base-v2}; max $50$ relations and $k=100$ rows per call \\
Passage retrieval environment & Wiki-18 dump ($21{,}015{,}324$ passages), \texttt{E5-base-v2} + FAISS flat index, top-$3$ passages per call \\
\secbandit{} bandit parameters & 4 structural arms; EMA $\alpha = 0.5$; Boltzmann temperature $\eta = 0.4$; $Q_k(0) = 0.5$ \\
Numerical precision \& hardware & \texttt{bf16} on $8\times\text{NVIDIA A100}$ GPUs \\
Evaluation decoding & Greedy decoding ($T=0$) across all nineteen benchmarks \\
\bottomrule
\end{tabular}}
\end{table}

\FloatBarrier
\section{Complete Attribution Grids Across All Nineteen Benchmarks}
\label{app:sets}

Tables~\ref{tab:abl-kgqa}--\ref{tab:abl-reasoning} provide the complete nineteen-benchmark evaluation grids across all nine arms (Base model, two specialized baselines per family, and six controlled recipe variants of \rise{} on Qwen2.5-7B-Instruct) supporting Section~\ref{sec:main-results} and Section~\ref{sec:analysis}. Across all three capability families, the full grids confirm three consistent patterns: (1) combining the protocol cold start with on-policy RL (\rise{} w/ UNI) outperforms either stage in isolation (\rise{} w/o RL and \rise{} w/o SFT) on every family average; (2) reversing the curriculum order (\rise{} w/ H2E) consistently underperforms progressive ordering (\rise{} w/ E2H); and (3) closed-loop variance tracking (\rise{} w/ \secbandit{}) achieves the highest performance on complex multi-hop composition (WebQSP, CWQ, GrailQA, HotpotQA, MuSiQue, Bamboogle) and multi-step mathematical deduction (GSM8K, MATH500, AIME24, AMC23, K\&K Hard).

\begin{table}[!htbp]
\centering
\caption{Complete structured-knowledge grid for all recipe arms on Qwen2.5-7B-Instruct (RQ4), \%. Set-equality EM except SimpleQuestions and WebQuestions (any-item hit, $\dagger$) and MultiTQ (Hits@1). \textbf{Bold} marks the best arm of each column and \underline{underline} the second; the \rise{} w/ \secbandit{} row is shaded.}
\label{tab:abl-kgqa}
\tabbox{\begin{tabular}{@{}lccccccc@{}}
\toprule
Arm & WebQSP & CWQ & GrailQA & SimpleQ$^\dagger$ & WebQ$^\dagger$ & MultiTQ & Avg. \\
\midrule
Base model & 42.3 & 29.4 & 46.2 & 61.4 & 52.3 & 27.3 & 43.2 \\
\quad + PoG & 47.5 & 32.1 & 54.3 & 64.2 & 52.7 & 30.2 & 46.8 \\
\quad + KG-R1 & 82.4 & 65.6 & 62.4 & 68.8 & 64.5 & 33.0 & 62.8 \\
\quad \rise{} w/o RL & 59.6 & 57.1 & 62.2 & 68.3 & 58.5 & 28.4 & 55.7 \\
\quad \rise{} w/o SFT & 78.4 & 63.3 & 59.2 & 65.5 & 63.0 & 33.3 & 60.4 \\
\quad \rise{} w/ UNI & 84.1 & 79.4 & 64.6 & 70.4 & 65.6 & 33.6 & 66.3 \\
\quad \rise{} w/ H2E & 82.2 & 75.0 & 67.7 & 69.9 & 65.9 & 32.8 & 65.6 \\
\quad \rise{} w/ E2H & \second{90.7} & \second{83.2} & \second{84.8} & \best{72.6} & \best{67.5} & \second{34.3} & \second{72.2} \\
\riserow \quad \rise{} w/ \secbandit{} & \best{92.4} & \best{85.4} & \best{86.2} & \second{72.2} & \second{67.1} & \best{34.5} & \best{73.0} \\
\bottomrule
\end{tabular}}
\end{table}

\begin{table}[!htbp]
\centering
\caption{Complete cross-tool retrieval grid for all recipe arms on Qwen2.5-7B-Instruct (RQ4), \% (alias EM). \rise{} rows are evaluated zero-shot; Search-R1 and ZeroSearch are trained in-domain. \textbf{Bold} marks the best arm of each column and \underline{underline} the second.}
\label{tab:abl-searchqa}
\tabbox{\begin{tabular}{@{}lcccccccc@{}}
\toprule
Arm & NQ & TriviaQA & PopQA & HotpotQA & 2Wiki & MuSiQue & Bamboogle & Avg. \\
\midrule
Base model & 12.8 & 35.6 & 13.8 & 16.2 & 22.6 & 6.6 & 24.0 & 18.8 \\
\quad + Search-R1 & 42.4 & 63.4 & \second{51.6} & 32.8 & 33.2 & 17.4 & 26.4 & 38.2 \\
\quad + ZeroSearch & \second{43.6} & \second{65.2} & 48.8 & 34.6 & 35.2 & 18.4 & 38.4 & 40.6 \\
\quad \rise{} w/o RL & 32.3 & 55.3 & 37.3 & 37.3 & 36.2 & 16.3 & 38.4 & 36.1 \\
\quad \rise{} w/o SFT & 38.7 & 58.5 & 42.4 & 37.8 & 35.5 & 17.4 & 37.6 & 38.3 \\
\quad \rise{} w/ UNI & 40.1 & 62.7 & 47.6 & 38.5 & 37.6 & 18.0 & 42.4 & 41.0 \\
\quad \rise{} w/ H2E & 39.6 & 59.2 & 46.8 & 38.1 & 36.2 & 18.0 & 43.2 & 40.2 \\
\quad \rise{} w/ E2H & 42.8 & \best{65.4} & \best{51.8} & \second{39.2} & \best{39.8} & \second{18.8} & \second{44.8} & \best{43.2} \\
\riserow \quad \rise{} w/ \secbandit{} & \best{43.9} & 63.8 & 49.7 & \best{39.4} & \second{39.3} & \best{18.9} & \best{45.6} & \second{42.9} \\
\bottomrule
\end{tabular}}
\end{table}

\begin{table}[!htbp]
\centering
\caption{Complete tool-free general reasoning grid for all recipe arms on Qwen2.5-7B-Instruct (RQ4), \% (exact accuracy). \textbf{Bold} marks the best arm of each column and \underline{underline} the second.}
\label{tab:abl-reasoning}
\tabbox{\begin{tabular}{@{}lccccccc@{}}
\toprule
Arm & GSM8K & MATH500 & AIME24 & AMC23 & K\&K Easy & K\&K Hard & Avg. \\
\midrule
Base model & 30.2 & 42.6 & 3.3 & 37.5 & 41.5 & 18.5 & 28.9 \\
\quad + GraphWalker & 74.3 & 61.4 & 6.7 & 40.0 & 40.0 & 23.5 & 41.0 \\
\quad + SIE & 87.4 & 61.6 & \second{10.0} & \second{47.5} & \best{49.5} & 29.0 & 47.5 \\
\quad \rise{} w/o RL & 58.5 & 49.8 & \second{10.0} & 40.0 & 42.0 & 19.0 & 36.5 \\
\quad \rise{} w/o SFT & 78.2 & 58.4 & \second{10.0} & 40.0 & 46.5 & 27.5 & 43.4 \\
\quad \rise{} w/ UNI & 84.4 & 65.6 & \second{10.0} & 45.0 & 47.0 & 28.0 & 46.7 \\
\quad \rise{} w/ H2E & 85.3 & 64.0 & \second{10.0} & 45.0 & 47.0 & 27.5 & 46.5 \\
\quad \rise{} w/ E2H & \second{87.9} & \second{70.6} & \best{16.7} & \second{47.5} & \second{48.5} & \second{30.5} & \second{50.3} \\
\riserow \quad \rise{} w/ \secbandit{} & \best{88.2} & \best{73.4} & \best{16.7} & \best{52.5} & \second{48.5} & \best{31.5} & \best{51.8} \\
\bottomrule
\end{tabular}}
\end{table}

\FloatBarrier
\section{Multi-Turn Interaction Case Study}
\label{app:case}

Table~\ref{tab:case} presents a side-by-side multi-turn trajectory comparison between the base model and \rise{} on a multi-hop CWQ query. The example illustrates how the two policies diverge upon encountering a high-degree relation list and an error receipt: the base model issues a malformed unbracketed query and falls back to an ungrounded parametric guess, whereas \rise{} reformulates its query, traverses the intermediate entity node, and grounds the exact answer in the retrieved receipt.

\begin{table}[!htbp]
\centering
\caption{Side-by-side trajectory comparison on a multi-hop CWQ instance (\emph{``Which city was the director of Inception born in?''}, Topic entity: \texttt{[[Inception]]}). Rows marked $\dagger$ highlight recovery from relation-list fallback and syntax error receipts.}
\label{tab:case}
\tabbox{\begin{tabular}{@{}>{\raggedright\arraybackslash}p{0.47\linewidth}>{\raggedright\arraybackslash}p{0.47\linewidth}@{}}
\toprule
\textbf{Base Model (Qwen2.5-7B-Instruct)} & \textbf{\rise{} w/ \secbandit{} (Ours)} \\
\midrule
\textbf{Turn 1:} \texttt{<think>} I need to find the birthplace of the director of Inception. \texttt{</think>} \texttt{<search>} Where was the director of [[Inception]] born? \texttt{</search>} &
\textbf{Turn 1:} \texttt{<think>} First, I should find who directed the film [[Inception]]. \texttt{</think>} \texttt{<search>} Who directed [[Inception]]? \texttt{</search>} \\
\addlinespace[3pt]
\textbf{Receipt 1$^\dagger$:} \texttt{Resolved: Inception (m.0661ql3) | intent=relation\_list} \newline \texttt{No single relation matched; available relations (12 of 41): film.film.directed\_by, film.film.starring, film.film.country, ... Calls remaining: 15} &
\textbf{Receipt 1:} \texttt{Resolved: Inception (m.0661ql3) | intent=relation\_hop} \newline \texttt{Results (1 of 1): Inception | film.film.directed\_by | Christopher Nolan. Calls remaining: 15} \\
\addlinespace[3pt]
\textbf{Turn 2:} \texttt{<think>} The graph did not directly return the city. Let me search for Christopher Nolan's birth city without brackets. \texttt{</think>} \texttt{<search>} birth city of Christopher Nolan \texttt{</search>} &
\textbf{Turn 2:} \texttt{<think>} The director is [[Christopher Nolan]]. Now I need to query his place of birth. \texttt{</think>} \texttt{<search>} What is the place of birth of [[Christopher Nolan]]? \texttt{</search>} \\
\addlinespace[3pt]
\textbf{Receipt 2$^\dagger$:} \texttt{Error [E\_NO\_ENTITY]: Query must reference at least one entity in double square brackets [[...]]. Calls remaining: 14} &
\textbf{Receipt 2:} \texttt{Resolved: Christopher Nolan (m.0bxtg) | intent=relation\_hop} \newline \texttt{Results (1 of 1): Christopher Nolan | people.person.place\_of\_birth | London. Calls remaining: 14} \\
\addlinespace[3pt]
\textbf{Turn 3:} \texttt{<think>} I know Christopher Nolan is British-American and lived in Chicago. \texttt{</think>} \texttt{<answer>} Chicago \texttt{</answer>} \textbf{(Ungrounded \& Wrong)} &
\textbf{Turn 3:} \texttt{<think>} The receipt confirms [[Christopher Nolan]] was born in London. \texttt{</think>} \texttt{<answer>} London \texttt{</answer>} \textbf{(Grounded \& Correct)} \\
\bottomrule
\end{tabular}}
\end{table}

\FloatBarrier
\section{Proof of Proposition 1 and Two-Stage Training Algorithm}
\label{app:proof}

\subsection{Proof of Proposition~\ref{prop:variance}}
\begin{proof}
If $r(\tau_i) = c$ for all $i \in \{1, \dots, G\}$, then $\bar{r} = c$ and $\hat{\sigma} = 0$, whence $\hat{A}_i = 0$ by definition and every gradient term in Eq.~(\ref{eq:grpo-objective}) vanishes. Under $r(\tau) - 1 \stackrel{\text{i.i.d.}}{\sim} \text{Bernoulli}(p)$, the group rewards are constant if and only if all $G$ rollouts succeed (probability $p^G$) or all $G$ fail (probability $(1-p)^G$), giving Eq.~(\ref{eq:nonzero-var-prob}). Differentiating $f(p) = 1 - p^G - (1-p)^G$ yields $f'(p) = G\big((1-p)^{G-1} - p^{G-1}\big)$, which vanishes on $(0,1)$ uniquely at $p = 0.5$, with $f''(0.5) < 0$ for all $G \ge 2$. Since $f(p)$ and $\sigma_{\text{Bern}}(p) = \sqrt{p(1-p)}$ are both symmetric about $p = 0.5$ and strictly increasing on $[0, 0.5]$, $f$ is a strictly comonotone function of $\sigma_{\text{Bern}}$.
\end{proof}

\subsection{Two-Stage Training Algorithm}
\label{app:algorithm}

Algorithm~\ref{alg:training} summarizes the complete two-stage training pipeline of \rise{}, combining protocol-preserving cold start on $\mathcal{Q}_{\text{cold}}$ (Stage 1) with observation-masked multi-turn GRPO and \secbandit{} learning-frontier scheduling on $\mathcal{Q}_{\text{RL}}$ (Stage 2).

\newlength{\algtextwidth}
\setlength{\algtextwidth}{\dimexpr\linewidth-2.5em\relax}

\begin{algorithm}[!htbp]
\caption{Two-Stage Training Pipeline of \rise{} with Self-Evolving Curriculum (\secbandit{}) Bandit}
\label{alg:training}
\begin{tabular}{@{}r@{\hspace{0.8em}}p{\algtextwidth}@{}}
\toprule
\multicolumn{2}{@{}p{\linewidth}@{}}{\textbf{Input:} base policy $\pi_{\theta_0}$; full graph $\mathcal{G}$ with typed adapter $\Phi_{\mathcal{G}}$; disjoint pools $\mathcal{Q}_{\text{cold}}$ ($6{,}000$ questions) and $\mathcal{Q}_{\text{RL}} = \bigcup_{k=1}^{4} \mathcal{Q}_{\text{RL}}^{(k)}$ ($28{,}215$ questions); call budget $C_{\max}=16$; group size $G=8$.} \\
\midrule
\multicolumn{2}{@{}p{\linewidth}@{}}{\textbf{Stage 1: Protocol-Preserving Cold Start}} \\
1: & \hangindent=1.15em\hangafter=1 Collect teacher rollouts inside $\Phi_{\mathcal{G}}$ on $\mathcal{Q}_{\text{cold}}$ and accept $\mathcal{D}_{\text{SFT}}$ by Eq.~(\ref{eq:sft-corpus}), i.e.\ $g(\tau)=1$, set-equality EM, $\varphi(\tau)=1.0$, and $z_t \neq \emptyset$ for every turn. \\
2: & \hangindent=1.15em\hangafter=1 Fine-tune $\pi_{\theta_0}$ on $\mathcal{D}_{\text{SFT}}$ to obtain the protocol-aligned checkpoint $\pi_{\text{ref}}$; set $\theta \leftarrow \theta_{\text{ref}}$ and $Q_k(0) \leftarrow 0.5$ for all $k \in \{1,2,3,4\}$. \\
\midrule
\multicolumn{2}{@{}p{\linewidth}@{}}{\textbf{Stage 2: On-Policy Multi-Turn GRPO with Learning-Frontier Scheduling}} \\
3: & \hangindent=1.15em\hangafter=1 \textbf{for} step $s = 1, \dots, S_{\text{epoch}}$ ($440$ steps) \textbf{do} \\
4: & \hangindent=2.30em\hangafter=1 \hspace*{1.15em}Sample arm probabilities $P_s(k) \propto \exp\!\big(Q_k(s-1)/\eta\big)$ by Eq.~(\ref{eq:sec-bandit}) and draw batch $\mathcal{B}_s = \bigcup_{k} \mathcal{B}_{s,k}$ of $\lvert\mathcal{B}_s\rvert = 64$ questions from $\mathcal{Q}_{\text{RL}}$. \\
5: & \hangindent=2.30em\hangafter=1 \hspace*{1.15em}\textbf{for each} question $q \in \mathcal{B}_s$ \textbf{in parallel do} \\
6: & \hangindent=3.45em\hangafter=1 \hspace*{2.30em}Roll out $G=8$ trajectories $\{\tau_i\}_{i=1}^{G} \sim \pi_{\theta_{\text{old}}}(\cdot \mid q)$ against $\Phi_{\mathcal{G}}$ under the call budget $C_{\max}$. \\
7: & \hangindent=3.45em\hangafter=1 \hspace*{2.30em}Score $r(\tau_i)$ by Eq.~(\ref{eq:reward}), form the group pass rate $p_q$, and set $\hat{A}_i = (r(\tau_i)-\bar{r})/\hat{\sigma}$, with $\hat{A}_i = 0$ whenever $\hat{\sigma} = 0$. \\
8: & \hangindent=2.30em\hangafter=1 \hspace*{1.15em}\textbf{end for} \\
9: & \hangindent=2.30em\hangafter=1 \hspace*{1.15em}Update $\theta$ by ascending the observation-masked gradient $\nabla_\theta \mathcal{J}_{\text{RISE}}(\theta)$ of Eq.~(\ref{eq:grpo-objective}). \\
10: & \hangindent=2.30em\hangafter=1 \hspace*{1.15em}For every active tier $k$, compute the frontier return $\bar{v}_k(s)$ over $\mathcal{B}_{s,k}$ and update $Q_k(s) \leftarrow (1-\alpha)Q_k(s-1) + \alpha \bar{v}_k(s)$. \\
11: & \hangindent=1.15em\hangafter=1 \textbf{end for} \\
\midrule
\multicolumn{2}{@{}p{\linewidth}@{}}{\textbf{Output:} interactive reasoning policy $\pi_{\theta}^{*}$.} \\
\bottomrule
\end{tabular}
\end{algorithm}

\FloatBarrier
\section{Discussion: Methodological Principles for Agentic RL}
\label{app:discussion}

Two broader methodological principles emerge from our study. First, \emph{reward modifications must be audited alongside answer-shape-invariant diagnostics}: any change to the outcome verifier that relaxes set equality or removes grounding gating ($g(\tau)$, Section~\ref{sec:reward-choice}) can produce apparent gains on exact-match metrics simply by altering the length or enumeration behavior of the policy's output. Enforcing multiplicative evidence grounding alongside strict set-equality verification is therefore essential for separating genuine search improvement from reward hacking.

Second, \emph{curriculum scheduling in group-normalized RL is fundamentally policy-indexed}: as established in Proposition~\ref{prop:variance}, a question contributes non-zero gradient if and only if its rollout group exhibits positive reward variance ($\hat{\sigma} > 0$). Because the competence frontier ($p(q;\theta) \approx 0.5$) shifts continuously as the policy updates and varies across model capacities, reversing the difficulty order (H2E) starves early updates of gradient, whereas progressive alignment (E2H) and online bandit tracking (\secbandit{}) keep batches anchored where the policy can actively learn.

\section{Limitations and Future Work}
\label{app:limitations}

We highlight three limitations that suggest directions for future work. First, under a finite call budget ($C_{\max}=16$), questions whose surface forms fail entity resolution or require complex semantic bridging beyond dense relation ranking remain unreachable; integrating adaptive entity-linking tools or hybrid graph-text retrieval within the same episode is a promising extension. Second, our curriculum experiments hold the training horizon fixed at one epoch ($440$ steps) to ensure clean, horizon-free attribution across all arms; exploring multi-epoch competence-frontier scheduling on larger heterogeneous tool ecosystems remains open. Third, every comparison here uses a single training run per arm, so scaling multi-seed training across larger model families ($>30\text{B}$ parameters) will further strengthen variance estimates.

\section{Implementation Audit and Taxonomy of Silent Failures}
\label{app:defects}

During the development of our multi-turn agentic RL pipeline, we conducted systematic turn-level audits that uncovered 23 silent implementation defects across three categories (summarized in Table~\ref{tab:defects}). Unlike traditional software bugs that raise runtime exceptions, silent failures in multi-turn RL permit training jobs to complete normally with smooth scalar reward curves while quietly degrading specific capabilities (such as single-turn termination or multi-answer grounding). Notably, our data-lineage audits also detected and removed $5$ test-split leakage items embedded inside upstream community training parquets (including $3$ instances carrying explicit \texttt{WebQTest-*} identifiers) and enforced strict train-only stratified partitioning between $\mathcal{Q}_{\text{cold}}$ and $\mathcal{Q}_{\text{RL}}$. We distill these into a reusable checklist to aid future agentic RL research.

\begin{table}[!htbp]
\centering
\caption{Taxonomy of 23 silent failure modes identified and resolved during our multi-turn RL pipeline audit.}
\label{tab:defects}
\tabbox{
\begin{tabular}{@{}lp{0.66\linewidth}@{}}
\toprule
Defect Category & Representative Silent Failure Modes \\
\midrule
Evaluation Instrument (3) & Non-greedy validation decoding ($T > 0$); tokenizer class mismatch upon checkpoint reload; subtle prompt-template divergence between rollout workers and standalone evaluators. \\
Data \& Cold-Start Lineage (8) & Cold-start question leakage from evaluation splits; upstream community training parquets containing test-split questions (\texttt{WebQTest-*}); SFT acceptance criteria looser than the RL reward gate $g(\tau)$; accepting teacher turns without preceding \texttt{<think>} blocks. \\
Rollout \& Trainer Config (12) & Binding the search call budget to the per-turn token cap; environment receipt tokens not masked in policy-gradient loss; missing stop-token strings in vLLM workers ($91\%$ wasted decoding); trainer-side right-truncation slicing \texttt{<information>} blocks mid-row. \\
\bottomrule
\end{tabular}}
\end{table}